\documentclass[10pt,journal,compsoc]{IEEEtran}
\usepackage{graphicx}
\usepackage{amsfonts,amssymb}
\usepackage{lipsum}
\usepackage{bm}
\usepackage{subeqnarray}
\usepackage{algorithm}
\usepackage{float}
\usepackage{algorithmic}
\usepackage{subfigure}
\usepackage{tabularx} 
\usepackage{amsmath}
\usepackage{booktabs }
\usepackage{amsthm}
\usepackage{hyperref}
\usepackage{url} 
\usepackage{cite}
\usepackage{multirow}
\graphicspath{{figures/}}
\usepackage[numbers]{natbib}
\begin{document}
\title{Differentially Private Distributed Online Learning }
\author{\normalsize Chencheng Li, \emph{Student Member, IEEE}, \normalsize Pan Zhou\dag, Gong Chen \emph{Member, IEEE} and Tao Jiang, \emph{Senior Member, IEEE}\IEEEcompsocitemizethanks{\IEEEcompsocthanksitem Chencheng Li, \dag Corresponding author of this paper, P. Zhou and T. Jiang are with the School
of Electronic Information and Communications, Huazhong University of
Science and Technology, Wuhan 430074, China.
E-mail: lichencheng@hust.edu.cn, panzhou@hust.edu.cn, tao.jiang@ieee.org
\IEEEcompsocthanksitem Gong Chen is with the School of Electrical
and Computer Engineering,  Georgia Tech.
E-mail: gong.chen@gatech.edu }
\thanks{Manuscript received XXXXX; revised XXXXX.}
}

\markboth{IEEE TRANSACTIONS ON KNOWLEDGE AND DATA ENGINEERING, VOL. X, NO. X, JANUARY 20XX}
{Li \MakeLowercase{\textit{et al.}}: Differentially Private Distributed Online Learning}

\IEEEtitleabstractindextext{%
\begin{abstract}In this paper, we propose a novel distributed online learning algorithm to handle massive data in Big Data era. Comparing to the typical centralized scenario, our proposed distributed online learning has multi-learners. Each learner optimizes its own learning parameter based on local data source and communicates timely with neighbors. We study the regret of the distributed online learning algorithm. However, communications among the learners may lead to privacy breaches. Thus, we use differential privacy to preserve the privacy of the learners, and study the influence of guaranteeing differential privacy on the regret of the algorithm. Furthermore, our online learning algorithm can be used to achieve fast convergence rates for offline learning algorithms in distributed scenarios. We demonstrate that the differentially private offline learning algorithm has high variance, but we can use mini-batch to improve the performance. The simulations show that our proposed theorems are correct and our differentially private distributed online learning algorithm is a general framework.
\end{abstract}

}
\maketitle
\IEEEdisplaynontitleabstractindextext
\IEEEpeerreviewmaketitle

\section{Introduction}
As the Internet develops rapidly, increasingly more information is put online. For example, in daily life, tens of millions of people on Facebook often share their photos on personal pages and post stories of life in the comments,  which  makes  Facebook process a large scale of data every second. Processing such a large scale of data in an efficient way is a challenging issue. In addition,  as an online interaction platform, Internet should offer people a real-time service. This makes Internet companies (e.g., Google, Facebook and YouTube) have to response and update their systems in real time. To provide better services, they need to learn and predict the user  behavior based on the past information of users. Hence, the notion ``online learning'' was introduced by researchers. In early stages, most online learning algorithms proceed in a centralized approach. However, as the data volume grows exponentially large in Big Data era, typical centralized online learning algorithms are no longer capable of  processing  such \emph{large-scale} and \emph{high-rate} online data. Besides, online data collection is inherently decentralized  because data sources are often widely distributed in different geographical locations. So it is much more natural to develop a distributed online learning algorithm (DOLA) to solve the problem.

During the learning process,  sharing information may leads to privacy breaches.  For instance, the hospitals in a city want to conduct a survey (can be regarded as a learning process) of the diseases that citizens are susceptible to. To protect the sensitive information of patients, the hospitals obviously can't release their cases of illness. Instead, each hospital just can share some limited information with other hospitals. However, different patient samples lead to different results. Through analyzing the results, the adversary is able to obtain some sensitive information about certain patients whose cases are only included in one hospital. Faced with this kind of privacy breach, the problem is how we can preserve the privacy of participants in the survey without significantly affecting the accuracy of the survey. To solve this class of problems, we urge to propose a privacy-preserving algorithm, which not only effectively processes distributed online learning, but also protects the privacy of the learners.

In this paper, we propose a differentially private distributed online learning algorithm with decentralized learners and data sources. The algorithm addresses two issues: 1) distributed online learning; 2) privacy-preserving guarantees. Specifically, we use \emph{distributed convex optimization} as the distributed online learning model, while use \emph{differential privacy} \citep{dwork2006differential} to protect the privacy.

Distributed convex optimization is considered as a \emph{consensus problem} \citep{olfati2007}.  To solve this problem, some related works \cite{ram2010distributed, nedic2009distributed,yuan2013convergence} have been done. These papers considered a multi-agents network system, where they studied distributed convex optimization for minimizing a sum of convex objective functions. For the convergence of  their algorithms, each agent updates the  iterates with usual convex optimization method and communicates the iterates to its neighbors. To achieve this goal, a time-variant communication  matrix is used to conduct the communications among the agents. The time-variant communication  matrix makes the distributed optimization algorithm converge faster and better  than the fixed one used in \cite{Yan2013}. For our work, the first issue is how the DOLA performs compared with the centralized algorithm. To this end, we use some results of the above works to compute the regret bounds of our DOLA.

 Differential privacy \citep{dwork2006differential} is  a popular privacy mechanism to preserve the privacy of the learners. A lot of progress has been made on differential privacy. This  mechanism prevents the adversary from gaining any meaningful information of any individuals. This privacy-preserving method is scalable for large and dynamic dataset. Specifically, it can provide the rigorous and quantitative  demonstrations for the risk of a privacy breach in statistical learning algorithms.  Many privacy-preserving algorithms \citep{jain2011differentially,chaudhuri2011,williams2010} have been proposed to use differential privacy to protect sensitive information in the centralized offline learning framework. However, in the distributed learning framework, there is seldom research effort.

Furthermore, our differentially private DOLA  can be used to achieve fast convergence rates for differentially private distributed offline learning algorithm based on \cite{kakade2009generalization}. Since the offline learning algorithm has access to all data, the technique of mini-batch \citep{song2013stochastic} is used to reduce the high variance of the differentially private offline learning algorithm. Motivated by \cite{kakade2009generalization} and \citep{song2013stochastic}, we try to obtain a good utility of the distributed offline learning algorithm while protect the privacy of the learners. More importantly, our differentially private distributed offline learning algorithm guarantees the same level of privacy as the DOLA with less random noise and achieves fast convergence rate.

Following are the main contributions of this paper:
 \begin{itemize}
 \item We present a DOLA (i.e., Algorithm 1), where each learner updates its learning  parameter based on local data source and exchanges information with  neighbors. We respectively obtain the classical regret bounds $O(\sqrt T )$ \cite{zinkevich2003} and $O(\log T)$  \cite{hazan2007} for convex and strongly convex objective functions for the algorithm.

 \item To protect the privacy of  learners, we make our DOLA guarantee $\epsilon$-differential privacy. Interestingly, we find that the private regret bounds has the same order of $O(\sqrt T )$ and $O(\log T)$  with the non-private ones, which indicates that guaranteeing differential privacy in the DOLA do not significantly hurt the original performance.

 \item We use the differentially private DOLA with good regret bounds to solve differentially private  distributed offline learning problems  (i.e., Algorithm 2) for the first time. We make Algorithm 2 have tighter utility guarantees than the existing state-of-the-art results while guarantee $\epsilon$-differential privacy.

 \item  We use mini-batch  to reduce  high variance of  the differentially private distributed offline learning algorithm and demonstrate that  the  algorithm using mini-batch guarantees the same level of privacy with less noise.
 \end{itemize}

The rest of the paper is organized as follows. Section 2 discusses some related works. Section 3 presents preliminaries for the formal  distributed online learning. Section 4 proposes the differentially private distributed online learning algorithm. We discuss the privacy analysis of our DOLA in Section 4.1 and discuss the regret bounds in Section 4.2. In Section 5, we discuss the application of the DOLA to the  differentially private distributed offline learning algorithm. Section 5.1 and 5.2 discuss the privacy and the regret respectively. In Section 6, we present simulation results of the proposed algorithms. Finally, Section 7 concludes the paper.

\section{Related Work}
Jain et al. \cite{jain2011differentially} studied the differentially private centralized online learning. They provided a generic differentially private framework for online algorithms. They showed that using their generic framework, Implicit Gradient Descent (IGD) and Generalized Infinitesimal Gradient Ascent (GIGA) can be transformed into differentially private online learning algorithms. Their work motivates our study on the differentially private online learning in distributed scenarios.

Recently, growing research effort has been devoted to distributed online learning. Yan et al. \cite{Yan2013} has proposed a DOLA to handle the decentralized data. A fixed network topology was used to conduct the communications among the learners in their system. They analyzed the regret bounds for convex and strongly convex functions respectively. Further, they studied the privacy-preserving  problem, and showed that the communication network made their algorithm have intrinsic privacy-preserving properties. Worse than differential privacy, their privacy-preserving method cannot protect the privacy of all learners absolutely. Because their privacy-preserving properties depended on the connectivity between two nodes, however,  all the  nodes cannot have the same connectivity in a fixed communication matrix. Besides, Huang et al. \cite{huang2015} is  closely related to our work. In their paper, they presented a differentially private distributed optimization algorithm. While guaranteed the convergence of the algorithm, they used differential privacy to protect the privacy of the agents. Finally, they observed that  to guarantee $\epsilon$-differential privacy,  their algorithm had the accuracy of the order of $O(\frac{1}{{{\epsilon ^2}}})$. Comparing to this accuracy, we obtain not only $O(\frac{1}{{{\epsilon ^2}}})$ rates for convex functions, but also $O(\frac{1}{{{\epsilon}}})$ rates for strongly convex functions, if our regret bounds of the differentially private DOLA are converted to convergence rates

The method to solve distributed online learning was pioneered in distributed optimization. Hazan has studied online convex optimization in his book \cite{hazan2015online}. They proposed that the framework of convex online learning is closely tied to statistical learning theory and convex optimization.  Duchi et al. \cite{duchi2012dual} developed an efficient algorithm for distributed optimization based on dual averaging of subgradients method. They demonstrated that their algorithm could work, even the communication matrix is random and not fixed. Nedic and Ozdaglar \cite{nedic2009distributed} considered a subgradient method for distributed convex optimization, where the functions are convex but not necessarily smooth. They demonstrated that a time-variant communication could ensure the convergence of the distributed optimization algorithm. Ram et al. \cite{ram2010distributed} tried to analyze the influence of stochastic subgradient errors on distributed convex optimization based on a time-variant network topology. They studied the convergence rate of their distributed optimization algorithm. Our work extends   the works of Nedic and Ozdaglar \cite{nedic2009distributed} and Ram et al. \cite{ram2010distributed}. All these papers have made great contributions to distributed convex optimization, but they did not consider the privacy-preserving problem.

As for the study of differential privacy, there has been much research effort being devoted to how differential privacy can be used in  existing learning algorithms. For example, Chaudhuri et al. \cite{chaudhuri2011} presented the output perturbation and objective perturbation ideas about differential privacy in empirical risk minimization (ERM) classification. They  achieved a good utility for ERM algorithm while guaranteed $\epsilon$-differential privacy. Rajkumar and Agarwal \cite{rajkumar2012} extended differentially private ERM classification \cite{chaudhuri2011} to differentially private ERM multiparty classification. More importantly, they analyzed the sequential and parallel composability problems while the algorithm guaranteed $\epsilon$-differential privacy. Bassily et al. \cite{Bassily2014} proposed more efficient algorithms and tighter error bounds for ERM classification on the basis of \cite{chaudhuri2011}.

Some papers have discussed the application of online learning with good regret to offline learning. Kakade and Tewari \cite{kakade2009generalization} proposed some properties of online learning algorithms if the loss function is Lipschitz and strongly convex. They found that recent online algorithms with logarithmic  regret guarantees could help to achieve fast convergence rates for the excess risk with high probability. Subsequently, Jain et al. \cite{jain2011differentially} use the results in \cite{kakade2009generalization} to analyze the utility of differentially private offline learning algorithms. 

\section{Preliminaries}
\textbf{Notation:} Upper case letters (e.g., $A$ or $W$) denote matrices or data sets, while lower case letters (e.g., $a$ or $w$) denote elements of matrices or column vectors. For instance, we denote the $i$-th learner’s parameter vector at time $t$ by $w_t^i$. $w[j]$ denotes the $j$-th component of a vector $w$ of length $N$. $a_ {ij}$ denotes the $(i,j)$-th element of $A$. Unless special remark, $\left\|  \cdot  \right\|$ denotes the Euclidean norm $\left\| w \right\|: = \sqrt {\sum\nolimits_i {w[i]^2} }$  and   $\left\langle { \cdot , \cdot } \right\rangle$ denotes the inner product $\left\langle {x,y} \right\rangle  = {x^{\rm T}}y$. ${\alpha _t}$ denotes the stepsize.

\textbf{Centralized Online Learning:} Given the information of the correct results to previous predictions, online learning aims at making a sequence of predictions. Online learning algorithms proceed in rounds.  At round $t$, the learner gets a question ${x_t}$, taken from a convex set $X$ and should give an answer denoted by $p_t$ to this question. Finally, the correct answer $y_t$ is given to be compared with $p_t$. Specifically, in online regression problems, ${x_t}$ denotes a vector of features, then ${p_t} \leftarrow \left\langle {{w_t},{x_t}} \right\rangle$ is a sequence of linear predictions, and comparing $p_t$ with $y_t$ leads to the loss function $\ell \left( {{w_t},{x_t},{y_t}} \right)$ (e.g., $\ell \left( {{w_t},{x_t},{y_t}} \right) = \left| {\left\langle {{w},{x_t}} \right\rangle  - {y_t}} \right|$). We  let ${f_t}(w): = \ell (w,{x_t},{y_t})$, which is obviously a convex function. According to the definition of online learning regret, the goal of online learning model is to minimize the function:
\begin{eqnarray}
{R_C} = \sum\limits_{t = 1}^T {{f_t}({w_t}) - \mathop {\min }\limits_{w \in W} } \sum\limits_{t = 1}^T {{f_t}(w)}, 
\end{eqnarray}
 where $W \subseteq {\mathbb{R}^n}$.

 In this paper, distributed online learning model is developed on the basis of the above description. 

\textbf{Distributed Convex Optimization:} Besides basic assumptions for datasets and objective functions, how conducting the communications among the distributed learners is critical to solve the distributed convex optimization problem in our work. Since the learners exchange information with neighbors while they update local parameters with subgradients, a time-variant  $m$-by-$m$ doubly stochastic matrix ${A_t}$ is proposed to conduct the communications.  ${A_t}$ has a few properties: 1) all elements of ${A_t}$ are non-negative and the sum of each row or column is one; 2) ${a_{ij}}(t)>0$ means there exists a communication between the $i$-th and $j$-th learners at round $t$, while ${a_{ij}}(t)=0$  means non-communication between them; 3) there exists a constant $\eta$, $0<\eta<1,$ such that ${a_{ij}}(t)>0$ implies that ${a_{ij}}(t)>\eta$. 

For distributed convex optimization, two assumptions must be made. First, we make the following assumption on the dataset $W$ and the cost functions $f_t^i$.

\textbf{Assumption 1.} The set $W$ and the cost functions $f_t^i$ are such that
\begin{itemize}
\item[(1)] The set $W$ is closed and convex subset of $\mathbb{R}^n$. Let $R \buildrel \Delta \over = \mathop {\sup }\limits_{x,y \in W} \left\| {x - y} \right\|$ denote the diameter of $W$.
\item[(2)] The cost functions $f_t^i$ are \emph{strongly convex} with modulus $\lambda  \ge 0$. For all $x,y \in W$, we have
\begin{eqnarray}
\left\langle {\nabla f_t^i,y - x} \right\rangle  \le f_t^i(y) - f_t^i(x) - \frac{\lambda }{2}{\left\| {y - x} \right\|^2}.
\end{eqnarray}
\item[(3)] The subgradients of $f_t^i$ are uniformly bounded, i.e., there exists $L > 0$ , for all $x \in W$, we have
\begin{eqnarray}\left\| {\nabla f_t^i(x)} \right\| \le L.\end{eqnarray}
\end{itemize}

Assumption (1) guarantees that there exists an optimal solution in our algorithm. Assumptions (2) and (3) help us analyze the convergence of our algorithm.

To recall, the learners communicate with neighbors based on the matrix of $A_t$. Each learner directly or indirectly influences  other learners. For a clear description, we denote the communication graph for a learner $i$ at round $t$ by 
\begin{eqnarray}{\mathcal{G}(t)_i} = \{ (i,j):{a_{ij}}(t) > 0\},\end{eqnarray}
 where \[{a_{ij}}(t) \in {A_{t}}.\]In our algorithm, each learner computes a weighted average \cite{ram2010distributed} of the $m$ learners' parameters. For the convergence of the DOLA, the weighted average should make each learner have ``equal" influence on other learners in  long rounds.  Then, we make the following assumption about the properties of $A_t$.

\textbf{Assumption 2.} For an arbitrary learner $i$, there exist a minimal scalar $\eta$, $0 < \eta  < 1$, and a scalar $N$ such that
\begin{itemize}
\item[(1)]${a_{ij}}(t) > 0 $ for $(i,j) \in {C_G}(t + 1)$,
\item[(2)] $\sum\nolimits_{j = 1}^m {{a_{ij}}(t)}  = 1$ and $\sum\nolimits_{i = 1}^m {{a_{ij}}(t)}  = 1$,
\item[(3)] ${a_{ij}}(t) > 0$ implies that  ${a_{ij}}(t + 1) \ge \eta $,
\item[(4)]  The graph ${ \cup _{k = 1,...N}}\mathcal{G}(t + k)_i$ is strongly connected for all $k$.
\end{itemize}

Here, Assumptions (1) and (2) state that each learner computes a weighted average of the parameters shown in Algorithm 1. Assumption (3) ensures that the influences among the learners are significant. Assumptions (2) and (4) ensure that the $m$ learners are equally influential in a long run. Assumption 2 is crucial to minimize the regret bounds in distributed scenarios.

\textbf{Differential Privacy:} Dwork \cite{dwork2006differential} proposed the definition of differential privacy for the first time. Differential privacy makes a data miner be able to release some statistic of its database without revealing sensitive information about a particular value itself. In this paper, we use differential privacy to protect the privacy of learners and  give the following definition.

\textbf{Definition 1.} Let $\mathcal{A}$ denote our differentially private DOLA. Let $\mathcal{X}=\left\langle {x_1^i,x_2^i,...,x_T^i} \right\rangle $ be a sequence of questions taken from an arbitrary learner's  local data source. Let $\mathcal{W} = \left\langle {w_1^i,w_2^i,...,w_T^i} \right\rangle $ be a sequence of $T$ outputs of  the learner and $\mathcal{W} =\mathcal{A}(\mathcal{X})$. Then, our algorithm $\mathcal{A}$ is $\epsilon$-differentially private if given any two adjacent question sequences $\mathcal{X}$ and $\mathcal{{X'}}$ that differ in one question entry, the following holds: 
 \begin{eqnarray}\Pr \left[ {\mathcal{A\left( X \right)} \in W} \right] \le {e^\epsilon}\Pr \left[ {\mathcal{A\left( {X'} \right)} \in W} \right].\end{eqnarray} 

This  inequality guarantees that whether or not an individual participates in the database, it will not make any significant difference on the output of our algorithm, so the adversary is not able to gain useful information about the individual.

\section{Differentially Private Distributed Online Learning }
For differentially private distributed online learning, we assume to have a system of $m$ online learners,  each of them has the independent learning ability. The $i$-th learner updates its local parameter $w_t^i$ based on its local data points $\left( {x_t^i,y_t^i} \right)$ with $i \in \left[ {1,m} \right]$. The learner makes the prediction $\left\langle {w_t^i,x_t^i} \right\rangle$ at round $t$ , then the loss function $f_t^i(w): = \ell (w,x_t^i,y_t^i)$ is obtained. Even though the $m$ learners are distributed, each learner exchanges information with neighbors. Based on the time-variant matrix $A_t$, the learners communicate with different sets of their neighbors at different rounds, which makes them indirectly influenced by other data sources. Specifically, for a learner $i$, at each round $t$,  it  first gets the exchanged parameters and computes the weighted average of them, then updates the local parameter $w_t^i$ with respect to the weighted average $b_t^i$ and the subgradient $g_t^i$, finally broadcasts the new local parameter added with a random noise to its neighbors $\mathcal{G}(t)_i$. We summarize the algorithm in Algorithm 1.

Before we discuss the privacy and utility of Algorithm 1, the regret in distributed setting is given in the following definition.

\textbf{Definition 2.} In an online learning algorithm, we assume to have $m$ learners using local data sources. Each learner updates its parameter through a weighted average of the received parameters. Then, we measure the regret of the algorithm as
\begin{eqnarray}
 {R_D} = \sum\limits_{t = 1}^T {\sum\limits_{i = 1}^m {f_t^i} } (w_t^j) - \mathop {\min }\limits_{w \in W} \sum\limits_{t = 1}^T {\sum\limits_{i = 1}^m {f_t^i} } (w).
\end{eqnarray}

Obviously, ${{f_t}({w_t})}$  in (1) is changed to the sum of $m$ learners' loss function $\sum\nolimits_{i = 1}^m {f_t^i} (w_t^j)$ in (6). In centralized online learning algorithm, $N$ data points need $T=N$ rounds to be finished, while the distributed algorithm can handles $m \times N$ data points over the same time period. Notice that ${R_D}$ is computed with respect to an arbitrary learner's parameter $w_t^j$ \citep{Yan2013}. This states that single one learner can measure the regret of the whole system based on its local parameter, even though the learner do not handle all data in the system.

 Next, we analyze the privacy of Algorithm 1 in Section 4.1 and give the regret bounds in Section 4.2.
\begin{algorithm}
\caption{Differentially Private Distributed Online Learning}
\begin{algorithmic}[1]
\STATE \textbf{Input}: Cost functions $f_t^i(w ): = \ell (w,x_t^i,y_t^i)$, $i \in [1,m]$ and $t \in [0,T]$  ; initial points $w _0^1,...,w _0^m$; double stochastic matrix ${A_t} = (a_{ij}(t)) \in {R^{m \times m}}$; maximum iterations $T$.
\FOR {$t = 0,...,T$ }
 \FOR{each learner $i = 1,...,m$}
\STATE  $b_t^i = \sum\limits_{j = 1}^m {{a_{ij}}(t + 1)(w_t^j + \sigma _t^j)} $, where $\sigma _t^j$ is a Laplace noise vector in $\mathbb{R}^n$
\STATE  $g_t^i \leftarrow \nabla f_t^i(b_t^i)$
\STATE  $w_{t + 1}^i = {\mathop{\rm Pro}\nolimits} [b_t^i - {\alpha _{t + 1}}\cdot g_t^i]$ \\
(Projection onto $W$)
\STATE broadcast the output $(w_{t + 1}^i + \sigma _{t + 1}^i)$ to $\mathcal{G}(t)_i$
\ENDFOR
\ENDFOR
\end{algorithmic}
\end{algorithm}

\subsection{Privacy Analysis} 
As explained previously, exchanging information may cause some privacy breaches, so we have to use differential privacy to protect the privacy.  In the view of Algorithm 1, all  learners exchange their weighted parameters with neighbors at each round. For preserving-privacy, every exchanged parameter should be made to guarantee differential privacy. To achieve this target,  a random noise is added to the parameter $w_t^i$ (see step 7 in Algorithm 1). This method to guarantee differential privacy is known as \emph{output perturbation} \cite{chaudhuri2011}. We have known where to add noise, next we study how much noise to be added. 

Differential privacy aims at weakening the significantly difference between $\mathcal A\left( X \right)$ and
 $\mathcal A\left( {X'}\right)$. Thus, to show differential privacy, we need to know that how ``sensitive'' the algorithm $\mathcal{A}$ is. Further, according to \cite{dwork2006differential},  the magnitude of the noise depends on the largest change that a single entry in data source could have on the output of Algorithm 1; this quantity is referred to as the \emph{sensitivity} of the algorithm.  Then, we  define the sensitivity of Algorithm 1 in the following definition.

\textbf{Definition 3 (Sensitivity).} Recall in Definition 1, for any $\mathcal{X}$ and $\mathcal{{X'}}$, which differ in exactly one entry, we define the  sensitivity of Algorithm 1 at $t$-th round as
\begin{align}
{\rm{S}}({\rm{t}}){\rm{ = }}\mathop {\sup }\limits_{{\cal X},{\cal X}'} {\left\| {{\cal A}\left( {\cal X} \right){\rm{ - }}{\cal A}\left( {{\cal X}'} \right)} \right\|_{\rm{1}}}.
\end{align}

The above norm is $L_1$-norm. According to the notion of sensitivity, we know that  higher sensitivity leads to more   noise if the algorithm guarantees the same level of privacy. By bounding the sensitivity $S(t)$, we determine the magnitude of the  random noise to guarantee $\epsilon$-differential privacy. We  compute the bound of $S(t)$ in the following lemma.

\textbf{Lemma 1.} Under Assumption 1, if the $L_1$-sensitivity of the algorithm  is computed as (7), we obtain
 \begin{equation}{\rm{S}}(t) \le 2{\alpha _t\sqrt n L},
 \end{equation}
where $n$ denotes the dimensionality of vectors.
\begin{proof} Recall in Definition 1, $\mathcal{X}$ and $\mathcal{{X'}}$ are any two data sets differing in one entry. $w_t^i$ is computed based on the data set $\mathcal{X}$ while ${w_t^i}'$  is computed based on the data set $\mathcal{{X'}}$. Certainly, we have $ \left\| {\mathcal{A\left( X \right)}- \mathcal{A\left( {X'} \right)}} \right\|_1=\left\| {w_t^i -{w_t^i}'} \right\|_1$.

 For datasets $\mathcal{X}$ and $\mathcal{{X'}}$ we have
\[w_t^i = {\mathop{\rm Pro}\nolimits} \left[ {b_{t-1}^i - {\alpha _t}g_{t-1}^i} \right] \ and\ \
{w_t^i}' = {\mathop{\rm Pro}\nolimits} \left[ {b_{t-1}^i - {\alpha _t}{g_{t-1}^i}'} \right].\]

Then, we have
\begin{small} 
\begin{align}
\notag\left\| {w_t^i -{w_t^i}'} \right\|_1& = \left\| {{\mathop{\rm Pro}\nolimits} \left[ {b_{t-1}^i - {\alpha _{t}}g_{t-1}^i} \right] - {\mathop{\rm Pro}\nolimits} \left[ {b_{t-1}^i - {\alpha _{t}}{g_{t-1}^i}'} \right]} \right\|_1\\
\notag& \le \left\| {(b_{t-1}^i - {\alpha _{t}}g_{t-1}^i) - (b_{t-1}^i - {\alpha _{t}}{g_{t-1}^i}')} \right\|_1\\
\notag& = {\alpha _{t}}\left\| {g_{t-1}^i - {g_{t-1}^i}'} \right\|_1\\
\notag& \le {\alpha _{t}}\left( {\left\| {g_{t-1}^i} \right\|_1 + \left\| {{g_{t-1}^i}'} \right\|_1} \right)\\
\notag& \le {\alpha _t}\sqrt n \left( {{{\left\| {g_{t - 1}^i} \right\|}_2} + {{\left\| {g_{t - 1}^i}' \right\|}_2}} \right)\\
& \le 2{\alpha _{t}}\sqrt nL.
\end{align}\end{small}
By Definition 3, we know  \begin{equation}{\rm{S}}(t) \le \left\| {w_t^i -{w_t^i}'} \right\|_1.
 \end{equation}
Hence, combining (9) and (10), we obtain (8).\end{proof}

 We next determine the magnitude of the added random noise due to (10).  In step 7 of Algorithm 1, we use $\sigma$ to denote the random noise.
 $\sigma \in {\mathbb{R}^n}$ is a Laplace random noise vector drawn independently according to the density function:
\begin{equation}
Lap\left( {x|\mu } \right) = \frac{1}{{2\mu }}\exp \left( { - \frac{{\left| x \right|}}{\mu }} \right),
\end{equation}
where $\mu  = {{S\left( t \right)} \mathord{\left/ {\vphantom {{S\left( t \right)}\epsilon}} \right. \kern-\nulldelimiterspace} \epsilon}$.  We let $Lap\left( \mu  \right)$ denote the Laplace distribution. (8) and (10) show that the magnitude of the added random noise depends on the sensitivity parameters: $\epsilon$, the stepsize $\alpha _t$, the dimensionality of vectors $n$,  and the bounded subgradient $L$.

\textbf{Lemma 2.} Under Assumption 1 and 2, at the $t$-th round, the $i$-th online learner's output of $\mathcal{A}$, $\widetilde w_t^i$, is $\epsilon$-differentially private.
\begin{proof}
Let $\widetilde w_t^i = w_t^i + \sigma _t^i$ and $\widetilde {w}{_t^i}' = w_t^i + \sigma _t^i$, then
by the definition of differential privacy (see Definition 1), $\widetilde w_t^i$ is $\epsilon$-differentially private if
\begin{equation}\Pr [\widetilde w_t^i \in W] \le {e^\epsilon}\Pr [\widetilde {w}{_t^i}' \in W].
\end{equation}
For $w \in W$, we obtain
\begin{align}
\notag\frac{{\Pr \left( {\widetilde w_t^i} \right)}}{{\Pr \left( {\widetilde {w}{_t^i}'} \right)}}& = \prod\limits_{j = 1}^n {\left( {\frac{{\exp \left( { - \frac{{\epsilon\left| {w_t^i[j] - w[j]} \right|}}{{S\left( t \right)}}} \right)}}{{\exp \left( { - \frac{{\epsilon\left| {{w_t^i}'[j] - w[j]} \right|}}{{S\left( t \right)}}} \right)}}} \right)} \\
\notag &= \prod\limits_{j = 1}^n {\exp \left( {\frac{{\epsilon\left( {\left| {{w_t^i}'[j] - w[j]} \right| - \left| {w_t^i[j] - w[j]} \right|} \right)}}{{S\left( t \right)}}} \right)} \\
\notag& \le \prod\limits_{j = 1}^n {\exp \left( {\frac{{\epsilon\left| {{w_t^i}'[j] - w_t^i[j]} \right|}}{{S\left( t \right)}}} \right)} \\
\notag &= \exp \left( {\frac{{\epsilon{{\left\| {{w_t^i}' - w_t^i} \right\|}_1}}}{{S\left( t \right)}}} \right)\\
& \le \exp \left( \epsilon \right),
\end{align}
where the first inequality follows from the triangle inequality, and the last  inequality follows from (10).
\end{proof}
McSherry \cite{mcsherry2009} has proposed that the privacy guarantee does not degrade across rounds as the samples used in the rounds are disjoint. In Algorithm 1, at each round,  each learner is given a question $x_t^i$, then makes the prediction $w_t^i$. Finally, given the correct answers $y_t^i$, each learner can obtain the loss functions $f_t^i\left( w \right): = \ell (w,x_t^i,y_t^i)$. In this process, we regard $\left( {x_t^i,y_t^i} \right)$ as a sample. During the $T$ rounds of Algorithm 1, these samples are disjoint. Therefore, as  Algorithm 1 runs, the privacy guarantee will not degrade. Then we obtain the following theorem.

\textbf{Theorem 1 (Parallel Composition).} On the basis of Definition 1 and 3, under Assumption 1 and Lemma 2, our DOLA (see Algorithm 1) is $\epsilon$-differentially private.

\begin{proof} This proof follows from the theorem 4 of \cite{mcsherry2009}. The probability of the output $\mathcal{W}$ (defined in Definition 1)  is
\begin{equation}\Pr \left[ {A\left( X \right) \in W} \right] = \prod\limits_{t = 1}^T {\Pr [A{{\left( X \right)}_t} \in W]}. \end{equation}

Using the definition of differential privacy for each output (see Lemma 2), we have
\begin{align}
\notag&\prod\limits_{t = 1}^T {\Pr [\mathcal{A}{{\mathcal{\left( X \right)}}_t} \in W]} \\
\notag&\quad \le \prod\limits_{t = 1}^T {\Pr [\mathcal{A}{{\left( \mathcal{{X'}} \right)}_t} \in W]}  \times \prod\limits_{t = 1}^T {\exp \left( {\epsilon \times \left| {{\mathcal{X}_t} \oplus {\mathcal{X}^\prime }_t} \right|} \right)} \\
&\quad \le \prod\limits_{t = 1}^T {\Pr [\mathcal{A}{{\left( \mathcal{{X'}} \right)}_t} \in W]}  \times \exp \left( {\epsilon \times \left| {\mathcal{X} \oplus {\mathcal{X}^\prime }} \right|} \right),
\end{align}
where $ \left|\mathcal{X} \oplus {\mathcal{X}^\prime} \right|$ denotes the different entry between $\mathcal{X}$ and $\mathcal{X}^\prime$.
\end{proof}
Intuitively, the above inequality states that the ultimate privacy guarantee is determined by the worst of the privacy guarantees, not the sum $T\epsilon$. 

Combining (8), (11) and Lemma 2, we find that if each round of Algorithm 1 has the privacy guarantee at the same level ($\epsilon$-differential privacy), the magnitude of the noise will decrease as Algorithm 1 runs. That is because  the magnitude of the  noise depends on the stepsize $\alpha _{t + 1}$, which decreases as the subgradient descends.

\subsection{Regret Analysis}
 The regret of online learning algorithm represents a  sum of  mistakes, which are made by the learners  during the learning and predicting process. That means if Algorithm 1 runs better and faster, the regret of our distributed online learning algorithm will be lower. In other words, faster convergence rate ensures that the $m$ learners make less mistakes and predict more accurately. Hence, we bound the regret $R_D$ through the convergence of $w_t^i$ in Algorithm 1. 

To analyze the convergence of $w_t^i$, we  consider the behavior of the time-variant matrix $A_t$. Let $A_t$ be the matrix with $\left( {i,j} \right)$-th equal to ${a_{ij}}\left( t \right)$ in Assumption 2. According to the assumption, $A_t$ is a doubly stochastic.
As mentioned previously, some related works have studied the matrix convergence of ${A_t}$. For simplicity, we use one of these results to obtain the following lemma. 

 \textbf{Lemma 3 (\citep{ram2010distributed}).}   We suppose that at each round $t$, the matrix ${A_t}$  satisfies the description in Assumption 2. Then, we have
\begin{itemize}
\item[(1)] \emph{$\mathop {\lim }\limits_{k \to \infty } \phi (k,s) = \frac{1}{m}e{e^T}$  for all $k,s \in Z$ with $k \ge s,$\\
\vskip 0.5 mm
where 
\begin{align} \phi (k,s) = A(k)A(k - 1)A \cdot  \cdot  \cdot A(s + 1).\end{align}}
\item[(2)] Further, the convergence is geometric and the rate of convergence is given by
\begin{align}\left| {\left[ {\phi {{(k,s)}_{ij}} - \frac{1}{m}} \right]} \right| \le \theta {\beta ^{k - s}},\end{align}
where\begin{align} \notag \theta  = {\left( {1 - \frac{\eta }{{4{m^2}}}} \right)^{ - 2}}\quad\quad \beta  = {\left( {1 - \frac{\eta }{{4{m^2}}}} \right)^{\frac{1}{N}}}.\end{align}

\end{itemize}

Lemma 3 will be repeatedly used in the proofs of the following lemmas. Next, we study the convergence of Algorithm 1 in details. We use subgradient descent  method to make $w_t^i$ move forward to the theoretically optimal solution. Based on this method, we know that $w_{t+1}^i$ is closer to the optimal solution than $w_t^i$. Besides, we also want to know the difference between two arbitrary learners, but computing the norms $\left\| {w_t^i - w_t^j} \right\|$ makes no sense. Alternatively, we study the behavior of $\left\| {{{\overline w }_t} - w_t^i} \right\|$, where for all $t$, ${\overline w _t}$  is defined by
 \begin{eqnarray}\ {\overline w _t} = \frac{1}{m}\sum\limits_{i = 1}^m {w_t^i} . \end{eqnarray}

In the following lemma, we give the bound of $\left\| {{{\overline w }_t} - w_t^i} \right\|$.

\textbf{Lemma 4.} Under Assumption 1 and 2, for all $i \in \{ 1,...,m\}$ and $t \in \{ 1,...,T\}$, we have
\begin{equation}\left\| {{{\overline w }_t} - w_t^i} \right\| \le mL\theta \sum\limits_{k = 1}^{t - 1} {{\beta ^{t - k}}} {\alpha _k} + \theta \sum\limits_{k = 1}^{t - 1} {{\beta ^{t - k}}} \sum\limits_{i = 1}^m {\left\| {\sigma _k^i} \right\|}  + 2{\alpha _t}L.\end{equation}

\begin{proof} For simplicity, we first study $\left\| {{{\overline w }_{t+1}} - w_{t+1}^i} \right\|$ instead. 

Define that\begin{eqnarray}d_{t + 1}^i = w_{t + 1}^i - b_t^i,\end{eqnarray} where $b_t^i$ is defined in step 4 of Algorithm 1. We next estimate the norm of $d_t^i$ for any $t$ and $i$. According to the famous  non-expansive property of the Euclidean projection onto a closed and convex $W$, for all $x \in W$,  we have
 \begin{eqnarray}\left\| {\rm Pro[x]} \right\| \le \left\| x \right\|. \end{eqnarray} 
Based on (20) and (21), using the definition of $b_t^i$ and $g_t^i$ in Algorithm 1, we obtain 
\begin{align}
\notag  \left\| {d_{t + 1}^i} \right\| &=\left\| {{\mathop{\rm Pro}\nolimits} [b_t^i - {\alpha _{t + 1}}g_t^i] - b_t^i} \right\|\\
\notag  &\le {\alpha _{t + 1}}\left\| {g_t^i} \right\|\\ 
 &\le {\alpha _{t + 1}}L .
\end{align}
 We use (3)  in the last step.

We  conduct the mathematical induction for (20) and use the  matrices $\phi (k,s)$ defined in (16). We then obtain
\begin{align}
\notag w_{t + 1}^i = & d_{t + 1}^i + \sum\limits_{k = 1}^t {\left( {\sum\limits_{j = 1}^m {{{[\phi (t + 1,k)]}_{ij}}d_k^j} } \right)} \\
& + \sum\limits_{k = 1}^t {\left( {\sum\limits_{j = 1}^m {{{[\phi (t + 1,k)]}_{ij}}\sigma _k^j} } \right)} .
\label{equation:empirical}
\end{align}
Using (18) and (20), we rewrite ${{{\overline w }_{t+1}}}$ as follows
 \begin{equation}
\begin{split}
{\overline w _{t+1}} &= \frac{1}{m}\left( {{\sum\limits_{i = 1}^m {b_t^i} } + \sum\limits_{i = 1}^m {d_{t + 1}^i} } \right)\\
 &= \frac{1}{m}\left( {\sum\limits_{j = 1}^m {\sum\limits_{i = 1}^m {{a_{ij}}(t + 1)(w_t^i + \sigma _t^i)} }  + \sum\limits_{i = 1}^m {d_{t + 1}^i} } \right)\\
&= \frac{1}{m}\left( {\sum\limits_{i = 1}^m {\left( {\sum\limits_{j = 1}^m {{a_{ij}}(t + 1)} } \right)} (w_t^i + \sigma _t^i) + \sum\limits_{i = 1}^m {d_{t + 1}^i} } \right).
\end{split} \end{equation}
According to Assumption 2, we know $\sum\nolimits_{j = 1}^m {{a_{ij}}(t + 1)}  = 1$,
then simplify ${{{\overline w }_{t+1}}}$ as
 \begin{align}
\notag{\overline w _{t + 1}}& = \frac{1}{m}\left( {\sum\limits_{i = 1}^m {(w_t^i + \sigma _t^i)}  + \sum\limits_{i = 1}^m {d_{t + 1}^i} } \right)\\ 
 &= {\overline w _t} + \frac{1}{m}\sum\limits_{i = 1}^m {(\sigma _t^i + d_{t + 1}^i} ).
\end{align}
Finally, we have
 \begin{equation}
{\overline w _{t + 1}} = \frac{1}{m}\sum\limits_{k = 1}^t {\sum\limits_{i = 1}^m {\sigma _k^i} }  + \frac{1}{m}\sum\limits_{k = 1}^{t + 1} {\sum\limits_{i = 1}^m {d_k^i} }  .
\end{equation}
Using (23) and (26), we obtain
\begin{align}
\notag\left\| {{{\overline w }_{t{\rm{ + }}1}} - w_{t{\rm{ + }}1}^i} \right\|&{\rm{ = }}\left\| {\frac{1}{m}\sum\limits_{k = 1}^t {\sum\limits_{i = 1}^m {\sigma _k^i} }  + \frac{1}{m}\sum\limits_{k = 1}^{t + 1} {\sum\limits_{i = 1}^m {d_k^i} } } \right.\\
\notag&\quad - \left( {d_{t + 1}^i + \sum\limits_{k = 1}^t {\left( {\sum\limits_{j = 1}^m {{{[\phi (t + 1,k)]}_{ij}}d_k^j} } \right)} } \right.\\
\notag& \quad+ \left. {\left. {\sum\limits_{k = 1}^t {\left( {\sum\limits_{j = 1}^m {{{[\phi (t + 1,k)]}_{ij}}\sigma _k^j} } \right)} } \right)} \right\| \\
\notag& = \left\| {\sum\limits_{k = 1}^t {\sum\limits_{i = 1}^m {\left( {\frac{1}{m} - {{\left[ {\phi \left( {t + 1,k} \right)} \right]}_{ij}}} \right)} } } \right.\left( {\sigma _k^i + d_k^i} \right)\\
 &\quad+ \left. {\left( {\frac{1}{m}\sum\limits_{i = 1}^m {d_{t + 1}^i}  - d_{t + 1}^i} \right)} \right\|.
\end{align}
According to the triangle inequality  in Euclidean geometry, we further have 
\begin{align}
\notag\left\| {{{\overline w }_{t{\rm{ + }}1}} - w_{t{\rm{ + }}1}^i} \right\| \le &\sum\limits_{k = 1}^t {\sum\limits_{i = 1}^m {\left| {\frac{1}{m} - {{\left[ {\phi \left( {t + 1,k} \right)} \right]}_{ij}}} \right|} } \left( {\left\| {\sigma _k^i} \right\| + \left\| {d_k^i} \right\|} \right)\\
& + \frac{1}{m}\sum\limits_{i = 1}^m {\left\| {d_{t + 1}^i} \right\|}  + \left\| {d_{t + 1}^i} \right\|.
\end{align}
Using the bound of $\left\| {d_{t + 1}^i} \right\|$ in (22) and (17) in Lemma 3, we have
\begin{align}
\notag \left\| {{{\overline w }_{t{\rm{ + }}1}} - w_{t{\rm{ + }}1}^i} \right\| \le& mL\theta \sum\limits_{k = 1}^t {{\beta ^{t + 1 - k}}} {\alpha _k}\\
& + \theta \sum\limits_{k = 1}^t {{\beta ^{t + 1 - k}}} \sum\limits_{i = 1}^m {\left\| {\sigma _k^i} \right\|}  + 2{\alpha _{t + 1}}L.
\end{align}
Finally, we obtain (18) based on (29)\end{proof}

Next we bound the distance ${\left\| {{{\overline w }_{t + 1}} - w} \right\|^2}$ for an arbitrary $w \in W$. This bound together with Lemma 4 helps to analyze the convergence of our algorithm.

In following Lemma 5, 6 and Theorem 2, we denote ${f_t} = \sum\nolimits_{i = 1}^m {f_t^i}$ for simplicity.

\textbf{Lemma 5.} Under Assumption 1 and 2, for any $w \in W$ and for all $t$,  we have 
\begin{align}
\notag\left\| {{{\overline w }_{t + 1}} - w} \right\|  \le& \left( {1 + 2{\alpha _{t + 1}}L + 2L} \right. + \frac{2}{m}\sum\limits_{i = 1}^m {\left\| {\sigma _t^i} \right\|} \\
\notag&\left. { -2 \lambda } \right)\left\| {{{\overline w }_t} - w} \right\| - \frac{2}{m}\left( {{f_t}\left( {{{\overline w }_t}} \right) - {f_t}\left( w \right)} \right)\\
\notag& + 4L\frac{1}{m}\sum\limits_{i = 1}^m {\left\| {{{\overline w }_t} - w_t^i} \right\|} \\
& + {\left\| {\frac{1}{m}\sum\limits_{i = 1}^m {\left( {\sigma _t^i + d_{t+1}^i} \right)} } \right\|^2}.
\end{align}

\begin{proof} For any $w \in W$ and all t,  we use (25) to have 
\begin{align}
\notag{\left\| {{{\overline w }_{t + 1}} - w} \right\|^2} &= \left\| {{{\overline w }_t} + \frac{1}{m}\sum\limits_{i = 1}^m {\left( {\sigma _t^i + d_{t + 1}^i} \right)}  - w} \right\|\\
\notag& = {\left\| {{{\overline w }_t} - w} \right\|^2} + {\left\| {\frac{1}{m}\sum\limits_{i = 1}^m {\left( {\sigma _t^i + d_{t + 1}^i} \right)} } \right\|^2}\\
&\quad\,\,+ 2\left\langle {\frac{1}{m}\sum\limits_{i = 1}^m {\left( {\sigma _t^i + d_{t + 1}^i} \right)} ,{{\overline w }_t} - w} \right\rangle .
\end{align}
Based on
\begin{align}
&\notag\left\| {{{\overline w }_{t + 1}} - w} \right\| - \left\| {{{\overline w }_t} - w} \right\|\\
 &\notag\le \left( {\left\| {{{\overline w }_{t + 1}} - w} \right\| - \left\| {{{\overline w }_t} - w} \right\|} \right)\left( {\left\| {{{\overline w }_{t + 1}} - w} \right\| + \left\| {{{\overline w }_t} - w} \right\|} \right)\\
& = {\left\| {{{\overline w }_{t + 1}} - w} \right\|^2} - {\left\| {{{\overline w }_t} - w} \right\|^2},
\end{align}
 we  can transform (32) to the following inequality:
\begin{align}
\notag\left\| {{{\overline w }_{t + 1}} - w} \right\| & \le \left\| {{{\overline w }_t} + \frac{1}{m}\sum\limits_{i = 1}^m {\left( {\sigma _t^i + d_{t + 1}^i} \right)}  - w} \right\|\\
\notag&=  \left\| {{{\overline w }_t} - w} \right\| + {\left\| {\frac{1}{m}\sum\limits_{i = 1}^m {\left( {\sigma _t^i + d_{t + 1}^i} \right)} } \right\|^2}\\
&\quad\,\,+  2\left\langle {\frac{1}{m}\sum\limits_{i = 1}^m {\left( {\sigma _t^i + d_{t + 1}^i} \right)} ,{{\overline w }_t} - w} \right\rangle .
\end{align}

Now we pay attention to 
\begin{align}
\notag
&2\left\langle {\frac{1}{m}\sum\limits_{i = 1}^m {\left( {\sigma _t^i + d_{t + 1}^i} \right)} ,{{\overline w }_t} - w} \right\rangle \\
& = -\frac{2}{m}\sum\limits_{i = 1}^m { \left\langle {g_t^i,{{\overline w }_t} - w} \right\rangle }  + \frac{2}{m}\sum\limits_{i = 1}^m {\left\langle {g_t^i + \sigma _t^i + d_{t + 1}^i,{{\overline w }_t} - w} \right\rangle } ] .
\end{align}

 First, we compute the inner product:
 \begin{equation}\notag -\frac{2}{m}\sum\limits_{i = 1}^m { \left\langle {g_t^i,{{\overline w }_t} - w} \right\rangle }.
\end{equation} 
Using (2) and (3) in Assumption 1, we first obtain
\begin{align}
\notag& - \left\langle {g_t^i,{{\overline w }_t} - w} \right\rangle\\
 \notag& =  - \left\langle {g_t^i,{{\overline w }_t} - w_t^i} \right\rangle  - \left\langle {g_t^i,w_t^i - w} \right\rangle \\
\notag& \le \left\| {g_t^i} \right\|\left\| {{{\overline w }_t} - w_t^i} \right\| + f_t^i(w) - f_t^i(w_t^i) - \lambda\left\| {w_t^i - w} \right\|\\
\notag &= \left\| {g_t^i} \right\|\left\| {{{\overline w }_t} - w_t^i} \right\| + f_t^i({\overline w _t}) - f_t^i(w_t^i) - \lambda\left\| {w_t^i - w} \right\|\\
\notag& \quad\,+ f_t^i(w) - f_t^i({\overline w _t})\\
\notag& \le \left\| {g_t^i} \right\|\left\| {{{\overline w }_t} - w_t^i} \right\| + \left\langle {\overline g _t^i,{{\overline w }_t} - w_t^i} \right\rangle  - \lambda \left\| {w_t^i - {{\overline w }_t}} \right\|\\
\notag&\quad\, - \lambda\left\| {w_t^i - w} \right\| + f_t^i(w) - f_t^i({\overline w _t})\\
\notag& \le \left( {\left\| {g_t^i} \right\| + \left\| {\overline g _t^i} \right\|} \right)\left\| {{{\overline w }_t} - w_t^i} \right\|\\
\notag& \quad\,- \lambda \left\| {{{\overline w }_t} - w} \right\| + f_t^i(w) - f_t^i({\overline w _t})\\
& \le 2L\left\| {{{\overline w }_t} - w} \right\| -  \lambda \left\| {{{\overline w }_t} - w} \right\| - \left( {f_t^i({{\overline w }_t}) - f_t^i(w)} \right).
\end{align}
 Adding up the above inequality over $i=1,...,m$,  we can have
\begin{align}
 \notag&- \frac{2}{m}\sum\limits_{i = 1}^m {\left\langle {g_t^i,{{\overline w }_t} - w} \right\rangle }\\
 \notag& \le \frac{{4L}}{m}\sum\limits_{i = 1}^m {\left\| {{{\overline w }_t} - w_t^i} \right\|}  -2 \lambda \left\| {{{\overline w }_t} - w} \right\|\\
& \quad\,- \frac{2}{m}\left( {f_t({{\overline w }_t}) - f_t(w)} \right).
\end{align}

Then, compute the other inner product:
\begin{align}
\notag&\frac{2}{m}\sum\limits_{i = 1}^m {\left\langle {g_t^i + \sigma _t^i + d_{t + 1}^i,{{\overline w }_t} - w} \right\rangle } \\
\notag& \le \frac{2}{m}\sum\limits_{i = 1}^m {\left\| {g_t^i + \sigma _t^i + d_{t + 1}^i} \right\|} \left\| {{{\overline w }_t} - w} \right\|\\
\notag& \le \frac{2}{m}\sum\limits_{i = 1}^m {\left( {\left\| {g_t^i} \right\| + \left\| {\sigma _t^i} \right\| + \left\| {d_{t + 1}^i} \right\|} \right)} \left\| {{{\overline w }_t} - w} \right\|\\
& \le \frac{2}{m}\sum\limits_{i = 1}^m {\left( {{\alpha _{t + 1}}L + L + \left\| {\sigma _t^i} \right\|} \right)} \left\| {{{\overline w }_t} - w} \right\|.
\end{align}
In the last inequality, we use (3) and (16).

Combing (33)-(37), we  complete the proof.
\end{proof}
 Based on Lemma 4 and 5, we give the general regret bound in the following lemma. For simplicity, we let ${f_t} = \sum\nolimits_{i = 1}^m {f_t^i}$.

\textbf{Lemma 6.} We let ${w^ * }$ denote the optimal solution computed in hindsight. The regret $R_D$ of Algorithm 1 is given by:
\begin{align}
\notag&\sum\limits_{t = 1}^T {\left[ {{f_t}(w_t^i) - {f_t}({w^ * })} \right]}\\
\notag& \le \left( {mRL + \frac{{3\beta \theta {m^2}{L^2}}}{{1 - \beta }} + \frac{{13}}{2}m{L^2}} \right)\sum\limits_{t = 1}^T {{\alpha _t}} \\
\notag&\quad\, + \left( {\frac{{3\beta \theta mL}}{{1 - \beta }} + \frac{{2L + 1}}{{2m}}} \right)\sum\limits_{t = 1}^T {\sum\limits_{i = 1}^m {\left\| {\sigma _t^i} \right\|} } \\
 &\quad\,+ \frac{{mR}}{2}.
\end{align}

\begin{proof} We use (30) in Lemma 5, which contains the term ${f_t}({\overline w _t}) - {f_t}(w)$, and set $w=w^*$. Then, we rearrange (30) to have
\begin{align}
\notag&{f_t}(w_t^i) - {f_t}({w^ * })\\
\notag& = {f_t}({\overline w _t}) - {f_t}({w^ * }) + {f_t}(w_t^i) - {f_t}({\overline w _t})\\
\notag& \le \frac{m}{2}(1 - 2\lambda  + 2{\alpha _{t + 1}}L + 2L + \frac{2}{m}\sum\limits_{i = 1}^m {\left\| {\sigma _t^i} \right\|} )\left\| {{{\overline w }_t} - {w^ * }} \right\|\\
\notag&\quad\, + 2L\sum\limits_{i = 1}^m {\left\| {{{\overline w }_t} - w_t^i} \right\|}  - \frac{m}{2}\left\| {{{\overline w }_{t + 1}} - {w^ * }} \right\|\\
& \quad\,+ \frac{m}{2}{\left\| {\frac{1}{m}\sum\limits_{i = 1}^m {\left( {\sigma _t^i + d_{t + 1}^i} \right)} } \right\|^2} + mL\left\| {{{\overline w }_t} - w_t^i} \right\|.
\end{align}

Plug in the bound of ${\left\| {{{\overline w }_t} - w_t^i} \right\|}$ in Lemma 4, we rewrite (39) as 
\begin{align}\notag
\notag&{f_t}(w_t^i) - {f_t}({w^ * })\\
\notag& \le \frac{m}{2}(1 - 2\lambda  + 2{\alpha _{t + 1}}L + 2L + \frac{2}{m}\sum\limits_{i = 1}^m {\left\| {\sigma _t^i} \right\|} )\left\| {{{\overline w }_t} - {w^ * }} \right\|\\
\notag&\quad\, - \frac{m}{2}\left\| {{{\overline w }_{t + 1}} - {w^ * }} \right\| + \frac{1}{{2m}}{\left\| {\sum\limits_{i = 1}^m {\left( {\sigma _t^i + d_{t + 1}^i} \right)} } \right\|^2} + 6{\alpha _t}m{L^2}\\
&\quad\, + 3\theta {m^2}{L^2}\sum\limits_{k = 1}^{t - 1} {{\beta ^{t - k}}{\alpha _k}}  + 3\theta mL\sum\limits_{k = 1}^{t - 1} {{\beta ^{t - k}}\sum\limits_{i = 1}^m {\left\| {\sigma _k^i} \right\|} } .
\end{align}

Summing up (40) over $t=1,...,T$, we have
\begin{small}
\begin{align}\notag
&\sum\limits_{t = 1}^T {\left[ {{f_t}({{\overline w }_t}) - {f_t}({w^ * })} \right]} \\
\notag &\le \underbrace {\frac{m}{2}\left[ {\sum\limits_{t = 1}^T {(1 - 2\lambda  + 2{\alpha _{t + 1}}L + 2L + \frac{2}{m}\sum\limits_{i = 1}^m {\left\| {\sigma _t^i} \right\|} )\left\| {{{\overline w }_t} - {w^ * }} \right\|} } \right.}_{ \in {S_1}}\\
\notag&\underbrace {\left. { - \sum\limits_{t = 1}^T {\left\| {{{\overline w }_{t + 1}} - {w^ * }} \right\|} } \right]}_{ \in {S_1}} + \underbrace {\sum\limits_{t = 1}^T {\left[ {\frac{1}{{2m}}{{\left\| {\sum\limits_{i = 1}^m {\left( {\sigma _t^i + d_{t + 1}^i} \right)} } \right\|}^2} + 6{\alpha _t}m{L^2}} \right]} }_{{S_2}}\\
&\underbrace { + 3\theta {m^2}{L^2}\sum\limits_{t = 1}^T {\sum\limits_{k = 1}^{t - 1} {{\beta ^{t - k}}{\alpha _k}} }  + 3\theta mL\sum\limits_{t = 1}^T {\sum\limits_{k = 1}^{t - 1} {{\beta ^{t - k}}\sum\limits_{i = 1}^m {\left\| {\sigma _k^i} \right\|} } } }_{{S_3}}.
\end{align}
\end{small}

Recall in Assumption 1, $R$ be the upper bound of the diameter of $W$ and ${\alpha _{t + 1}} < {\alpha _t}$, we compute (41) as follows
\begin{small}\begin{align}\notag
{S_1} &= \frac{m}{2}\sum\limits_{t = 2}^T {\left\| {{{\overline w }_t} - {w^ * }} \right\|\left( {2{\alpha _{t + 1}}L + 2L - 2\lambda  + \frac{2}{m}\sum\limits_{i = 1}^m {\left\| {\sigma _t^i} \right\|} } \right)} \\
\notag& \quad\,+ \frac{m}{2}\left\| {{{\overline w }_1} - {w^ * }} \right\|(1 + 2{\alpha _{t + 1}}L + 2L - 2\lambda  + \frac{2}{m}\sum\limits_{i = 1}^m {\left\| {\sigma _t^i} \right\|} )\\
\notag&\quad\, - \frac{m}{2}\left\| {{{\overline w }_{T + 1}} - {w^ * }} \right\|\\
\notag&  \le \frac{{mR}}{2}\sum\limits_{t = 1}^T {\left( {2{\alpha _t}L + \frac{2}{m}\sum\limits_{i = 1}^m {\left\| {\sigma _t^i} \right\|} } \right)}  + \frac{{mR}}{2} - mRT(\lambda  - L)\\
& \le R\sum\limits_{t = 1}^T {\left( {m{\alpha _t}L + \sum\limits_{i = 1}^m {\left\| {\sigma _t^i} \right\|} } \right)}  + \frac{{mR}}{2},
\end{align}\end{small}
\begin{small}\begin{align}\notag
{S_2}& = 3\theta {m^2}{L^2}\sum\limits_{t = 1}^T {\sum\limits_{k = 1}^{t - 1} {{\beta ^{t - k}}{\alpha _k}} }  + 3\theta mL\sum\limits_{t = 1}^T {\sum\limits_{k = 1}^{t - 1} {{\beta ^{t - k}}\sum\limits_{i = 1}^m {\left\| {\sigma _k^i} \right\|} } } \\
\notag& \le 3\theta {m^2}{L^2}\sum\limits_{t = 1}^T {{\alpha _t}\sum\limits_{k = 1}^T {{\beta ^k}} }  + 3\theta mL\sum\limits_{k = 1}^T {{\beta ^k}\sum\limits_{t = 1}^T {\sum\limits_{i = 1}^m {\left\| {\sigma _t^i} \right\|} } } \\
& \le \frac{{3\beta \theta {m^2}{L^2}}}{{1 - \beta }}\sum\limits_{t = 1}^T {{\alpha _t}}  + \frac{{3\beta \theta mL}}{{1 - \beta }}\sum\limits_{t = 1}^T {\sum\limits_{i = 1}^m {\left\| {\sigma _t^i} \right\|} } ,
\end{align}\end{small}
\begin{small}\begin{align}\notag
{S_3} &= 6m{L^2}\sum\limits_{t = 1}^T {{\alpha _t}}  + \frac{1}{{2m}}\sum\limits_{t = 1}^T {{{\left\| {\sum\limits_{i = 1}^m {\left( {\sigma _t^i + d_{t + 1}^i} \right)} } \right\|}^2}} \\
 \notag&\le 6m{L^2}\sum\limits_{t = 1}^T {{\alpha _t}}  + \frac{1}{{2m}}\sum\limits_{t = 1}^T {\left[ {\left( {2L + 1} \right)\sum\limits_{i = 1}^m {\left\| {\sigma _t^i} \right\|}  + {m^2}{L^2}{\alpha _t}} \right]} \\
& \le \frac{{13}}{2}m{L^2}\sum\limits_{t = 1}^T {{\alpha _t}}  + \frac{1}{{2m}}\sum\limits_{t = 1}^T {\left[ {\left( {2L + 1} \right)\sum\limits_{i = 1}^m {\left\| {\sigma _t^i} \right\|} } \right]}.
\end{align}\end{small}

Combining $S_1$, $S_2$ and $S_3$, we get (38).
\end{proof}

Lemma 6 gives the regret bound with respect to the stepsize $\alpha _t$ and the noise parameter $\sigma_t^i$. Further, we analyze the regret bounds for convex and strongly convex functions. Besides, we need to figure out the influence that the total noise have on the regret bounds.

\textbf{Theorem 2.} Based on Lemma 6, if $\lambda  > 0$ and we set ${\alpha _t} = \frac{1}{{\lambda t}}$, then the expected regret of our DOLA satisfies:
\begin{align}
\notag&\mathbb{E}\left[ {\sum\limits_{t = 1}^T {{f_t}(w_t^i)} } \right] - \sum\limits_{t = 1}^T {{f_t}({w^ * })} \\
\notag &\le \frac{{mL}}{\lambda }\left( {R + \frac{{3\beta \theta mL}}{{1 - \beta }} + \frac{{13}}{2}L} \right)\left( {1 + \log T} \right)\\
&\quad\,+ \left( {\frac{{3\beta \theta mL}}{{1 - \beta }} + \frac{{2L + 1}}{{2m}}} \right)\frac{{2\sqrt 2mnL }}{{\lambda\epsilon }}\left( {1 + \log T} \right) + \frac{{mR}}{2},
\end{align}
and if  $\lambda  = 0$  and set ${\alpha _t} = \frac{1}{{2\sqrt t }}$ then
\begin{align}
\notag&\mathbb{E}\left[ {\sum\limits_{t = 1}^T {{f_t}(w_t^i)} } \right] - \sum\limits_{t = 1}^T {{f_t}({w^ * })} \\
\notag& \le \frac{{mL}}{\lambda }\left( {R + \frac{{3\beta \theta mL}}{{1 - \beta }} + \frac{{13}}{2}L} \right)\left( {\sqrt T  - \frac{1}{2}} \right)\\
 &\quad\, + \left( {\frac{{3\beta \theta mL}}{{1 - \beta }} + \frac{{2L + 1}}{{2m}}} \right)\frac{{2\sqrt 2mnL }}{\epsilon }\left( {\sqrt T  - \frac{1}{2}} \right) + \frac{{mR}}{2}.
\end{align}

\begin{proof} First we consider ${\alpha _t} = \frac{1}{{\lambda t}}$, then
\begin{equation}\sum\limits_{t = 1}^T {{\alpha _t}}=\sum\limits_{t = 1}^T {\frac{1}{{\lambda t}}}  = \frac{1}{\lambda }\sum\limits_{t = 1}^T {\frac{1}{t}}  \le \frac{1}{\lambda }\left( {1 + \log T} \right).\end{equation}

Since $\sigma _t^i$ is drawn from $Lap\left( \mu  \right)$  and each component  of  the vector $\sigma _t^i$ is independent, we have 
\begin{small}\begin{align}
\notag\sum\limits_{t = 1}^T {\sum\limits_{i = 1}^m {\left\| {\sigma _t^i} \right\|} } & = m\sum\limits_{t = 1}^T {\left\| {{\sigma _t}} \right\|} \\
\notag &=\sum\limits_{t = 1}^T {m\sqrt {{{\left| {{\sigma _t}\left[ 1 \right]} \right|}^2}{\rm{ + }}...{\rm{ + }}{{\left| {{\sigma _t}\left[ n \right]} \right|}^2}} } \\
 &= m\sqrt n \sum\limits_{t = 1}^T {\sqrt {{{\left| {{\sigma _t}\left[ j \right]} \right|}^2}} } ,
\end{align}\end{small}
where ${\sigma _t\left[ j \right]}$ denotes an arbitrary component of the vector $\sigma _t$. Under the condition, ${\sigma _t}\left[ j \right] \sim Lap\left( \mu  \right)$, we have $\mathbb{E}\left[ {{{\left| {{\sigma _t}\left[ j \right]} \right|}^2}} \right] = 2{\mu ^2}$, then
\begin{align}
\notag E\left[ {\sum\limits_{t = 1}^T {\sum\limits_{i = 1}^m {\left\| {\sigma _t^i} \right\|} } } \right] &= E\left[ {m\sqrt n \sum\limits_{t = 1}^T {\sqrt {{{\left| {{\sigma _t}\left[ j \right]} \right|}^2}} } } \right]\\
\notag& = \sum\limits_{t = 1}^T {\mu m\sqrt {2n} } 
 = \sum\limits_{t = 1}^T {\frac{{S\left( t \right)m\sqrt {2n} }}{\epsilon}} \\
\notag& \le \frac{{2\sqrt 2mnL }}{\epsilon}\sum\limits_{t = 1}^T {{\alpha _t}} \\
 &\le \frac{{2\sqrt 2mnL }}{{\lambda\epsilon }}\left( {1 + \log T} \right).
\end{align}
The last inequality follows form (47).

Then, using (47) and (49), we get (45).

If $\lambda  = 0$,  and we set ${\alpha _t} = \frac{1}{{2\sqrt t }}$, we have
\begin{equation}\sum\limits_{t = 1}^T {{\alpha _t}}=\sum\limits_{t = 1}^T {\frac{1}{{2\sqrt t }}}  \le \sqrt T  - \frac{1}{2}.\end{equation}
Using (30), we rewrite (29) as 
\begin{align}
E\left[ {\sum\limits_{t = 1}^T {\sum\limits_{i = 1}^m {\left\| {\sigma _t^i} \right\|} } } \right] \le \frac{{2\sqrt 2mnL }}{a}\left( {\sqrt T  - \frac{1}{2}} \right).
\end{align}

Now, using (50) and (51) we get (46).
\end{proof}

As expected,  we  respectively  obtain the square root regret $O(\sqrt T)$ and the logarithmic regret $O(\log T )$  of  Algorithm 1 in Theorem 2. Intuitively, except for $T$, the regret bounds  are also  with respect to the size of distributed network $m$. More importantly, the total noise added to the outputs  has the magnitude of the same order of $O(\sqrt T )$ and $O(\log T)$. This means that guaranteeing differential privacy has no strong influence on the non-private DOLA. The reason why this happens is that 
the magnitude of the total noise is with respect to the stepsize $\alpha _t$ from (29). It has the similar form as the non-private regret.
Thus, the final regret bound with noise has the same order of non-private regret bound.

\section{Application to private distributed offline learning using mini-batch}

In Section 4, we proposed a differentially private DOLA with good regret bounds of $O(\sqrt T )$ and $O(\log T)$. Kakade and Tewari \cite{kakade2009generalization} and Jain et al. \cite{jain2011differentially}  have both proposed that online learning algorithms with good regret bounds can be used to achieve fast convergence rates for offline learning algorithms. Based on the analysis in  \cite{jain2011differentially},  we exploit this application in distributed scenarios. Before that, we first discuss the private distributed offline learning using mini-batch.

In distributed offline learning scenarios, we also assume that there are $m$ offline learners. Each learner can obtain the labelled examples $\left( {e.g.,\left( {x_1^i,y_1^i} \right),...\left( {x_n^i,y_n^i} \right)} \right)$ from its local data source. Differing from the distributed online learners, the offline learners have the data beforehand. Before we describe the distributed offline learning model,  we should  pay  attention to how the  centralized offline learning model works.

In a centralized offline learning model, the classical method of training such a model based on labelled data is  by optimizing the following problem:
\begin{equation}{w^ * } = \mathop {\arg \min }\limits_{w \in {R^n}} \frac{1}{n}\sum\limits_{k = 1}^n {\ell \left( {w,{x_k},{y_k}} \right)}  + \frac{\varphi }{2}{\left\| w \right\|^2},\end{equation}
where $\ell$ is a convex loss function. According to the different choices of $\ell$ in machine learning, we can obtain different data mining algorithms. For example, Support Vector Machine (SVM) algorithm comes from $\ell\left( {w,x,y} \right) = \max \left( {1 - y{w^T}x,0} \right)$ and Logistic Regression algorithm comes from $\ell\left( {w,x,y} \right) = \log \left( {1 + \exp \left( { - y{w^T}x} \right)} \right)$. For solving the problem in (52), stochastic gradient descent (SGD) (mentioned in \cite{song2013stochastic}) was proposed. SGD  updates the iterate at round $t$ as:
\begin{equation}{w_{t + 1}} = {w_t} - {\alpha _{t + 1}}\left( {\nabla \ell \left( {{w_t},{x_t},{y_t}} \right) + \varphi {w_t}} \right),\end{equation}
where this iterate is updated based on a single point $\left( {{x_t},{y_t}} \right)$ sampled randomly from the local data set. 

Next, based on the centralized offline learning model, we build the distributed offline learning model. In distributed model, each  learner updates its parameter with subgradient as (53) does. Meanwhile, each learner must exchange information with other learners. Hence, for distributed offline learning we update the iterate as:
\begin{equation}w_{t + 1}^i = \sum\limits_{j = 1}^m {{a_{ij}}(t + 1)} w_t^j - {\alpha _{t + 1}}\left( {g_t^i + \varphi w_t^i} \right).
\end{equation}

In offline leaning framework, all data are available beforehand. To handle such massive training points, we use SGD with mini-batch to update the iterate. Using mini-batch, we update the iterate at round $t$ on the basis of a subset $H_t$ of examples. This help us process  multiple sampled examples instead of a single one at each round. Under this model, our offline learning algorithm runs in a parallel and distributed method.  Based on mini-batch, we rewrite (54) as:
\begin{equation}w_{t + 1}^i = \sum\limits_{j = 1}^m {{a_{ij}}} \left( {t + 1} \right)\widetilde w_t^j - \frac{{{\alpha _{t + 1}}}}{h}\sum\limits_{({x_k},{y_k}) \in {H_t}} {g_k^i},\end{equation}
where $h$ denotes the number of examples included in $H_t$ and $\widetilde w_t^j$ is defined in Lemma 2. In (55), we compute an average of  subgradients of $h$ examples sampled i.i.d. from the local data source.

As with the DOLA, exchanging information also leads to a privacy breach in distributed offline learning. Hence, to protect the privacy, we  make our distributed offline learning algorithm guarantee $\epsilon$-differential privacy as well. The differentially private method used here  is the same with that used in Algorithm 1. Furthermore, mini-batch can weaken the influence of noise on regret bounds when the algorithm guarantees differential privacy. For example, Song et al. \cite{song2013stochastic} demonstrated that differentially private SGD algorithm updated with a single point has high variance and used mini-batch to reduce the variance. In this paper, we also use  mini-batch to achieve the same goal. 

 To conclude, we propose a private distributed offline learning algorithm using mini-batch. The algorithm is summarized in Algorithm 2.

\begin{algorithm}
\caption{Differentially Private Distributed Offline Learning Using Mini-Batch}
\begin{algorithmic}[1]
\STATE \textbf{Input}: Cost functions $f_t^i(w ): = \ell (w,x_t^i,y_t^i)$, $i \in [1,m]$ and $t \in [0,T]$  ; initial points $w _0^1,...,w _0^m$; double stochastic matrix ${A_t} = (a_{ij}(t)) \in {R^{m \times m}}$; maximum iterations $\frac{T}{h}$.
\FOR {$t = 0,...,\frac{T}{h}$ }
 \FOR{each learner $i = 1,...,m$}
\STATE  $b_t^i = \sum\limits_{j = 1}^m {{a_{ij}}(t + 1)(w_t^j + \sigma _t^j)} $, where $\sigma _t^j$ is a Laplace noise vector in $\mathbb{R}^n$
\STATE  $g_k^i \leftarrow \nabla f_k^i(b_t^i)$, which is computed based on examples $\left( {{x_k},{y_k}} \right) \in {H_t}$
\STATE  $w_{t + 1}^i = {\rm{Pro}}\left[ {b_t^i - {\alpha _{t + 1}}\left( {\varphi w_t^i + \frac{1}{h}\sum\limits_{\left( {{x_k},{y_k}} \right) \in {H_t}} {g_k^i} } \right)} \right]$ \\
(Projection onto $W$)
\STATE broadcast the output $(w_{t + 1}^i + \sigma _{t + 1}^i)$
\ENDFOR
\ENDFOR
\end{algorithmic}
\end{algorithm}

\subsection{Privacy analysis for  Algorithm 2}
 Algorithm 2 guarantees the same level of privacy as Algorithm 1 does. Differing from Algorithm 1, the step 6 in Algorithm 2 computes a average of subgradients. According to the analysis of the sensitivity in Section 4.1, we easily know that the  sensitivity of Algorithm 2 must be different from (8). Then, we compute new sensitivity of Algorithm 2 in the following  lemma. 

\textbf{ Lemma 7.} (Sensitivity of Algorithm 2) Under Assumption 1,  let  all definitions  made previously  be used here again, the $L_1$-sensitivity of  Algorithm 2  is
\begin{equation}S_2\left( t \right) \le \frac{{2{\alpha _t}\sqrt n L}}{h}.
\end{equation}

  We omit the proof of  Lemma 7, which follows along the lines of  Lemma 1.

Obviously, Lemma 7 demonstrates that except for the parameters in (6), the magnitude of the  sensitivity of Algorithm 2 is with respect to the batch size $h$. Comparing (56) with (8), we find that the  sensitivity of Algorithm 2 is smaller than that of Algorithm 1. (11) shows that lower sensitivity leads to less added noise. So Algorithm 2 can add less random noise to its output  while it guarantees the same level of privacy as Algorithm 1.

 To recall in Lemma 2, we also ensure that the output of Algorithm 2  guarantees $\epsilon$-Differential privacy at each round $t$. Then, we consider the following lemma.
 
\textbf{ Lemma 8.} At the $t$-th round, the $i$-th online learner's output of Algorithm 2 is $\epsilon$-differentially private.

The proof follows along the lines of Lemma 2, and is omitted.  

To recall, we use mini-batch to reduce the variance. We divide the dataset into batches  ${{{H}}_1},...,{H_t}$, which are disjoint  subsets. According to the theory of parallel composition \citep{mcsherry2009} in differential privacy, we know that the privacy guarantee does not degrade across rounds. Based on this observation, we can obtain the following theorem, which omits the proof.

\textbf{ Theorem 3.} Using Lemma 8 and the theory of parallel composition, Algorithm 2 is  $\epsilon$-differentially private.

\subsection{Utility  analysis for Algorithm 2}
As described, we next  use the regret bounds of  Algorithm 1 to achieve fast convergence rates for Algorithm 2 based on \cite{kakade2009generalization}. Note that the following Lemma 9 and 10 are proposed  to prepare for the final result, Theorem 4.

 For a clear description, we first consider the centralized offline learning. Let $X$  be the domain of samples $x_t$ and ${D_x}$ denotes a distribution over the domain $X$. Instead of  minimizing (1), we bound 
\begin{equation}
F\left( {\overline w } \right) - \mathop {\min }\limits_{w \in W} F\left( w \right),\end{equation}
 where $F\left( w \right) = E\left[ {f\left( {w,x,y} \right)} \right]$
 , $\left( {x,y} \right) \sim {D_x}$ and   $\overline w  = \frac{1}{T}\sum\nolimits_{t = 1}^T {{w_t}} $. Then, we obtain the centralized approximation error in the following lemma.

\textbf{ Lemma 9 (\citep{kakade2009generalization}). }Under Assumption 1, let $R_C$ be the regret (e.g., say $R_C \le \log T$) of centralized online learning algorithm. Then with probability $1 -4 \gamma \ln T$,
\begin{align}
\notag&F\left( {\overline w } \right) - F\left( {{w^ * }} \right)\\
 &\le \frac{{{R_C}}}{T} + 4\sqrt {\frac{{{L^2}\ln T}}{\lambda }} \frac{{\sqrt {{R_C}} }}{T} + \max \left\{ {\frac{{16{L^2}}}{\lambda },6} \right\}\frac{{\ln \left( {{1 \mathord{\left/
 {\vphantom {1 \gamma }} \right.
 \kern-\nulldelimiterspace} \gamma }} \right)}}{T},
\end{align}
where ${w^ * } \in \arg \mathop {\min }\limits_{w \in W} F\left( w \right)$.

Intuitively, Lemma 9 relates the online regret to the offline  convergence rate. But if we want to have the similar lemma when update the iterate as (55), we must know the  new online regret using mini-batch. Dekel et al. \cite{dekel2012optimal} demonstrated that the mini-batch update does not improve the regret but also not significantly hurt the update rule. Based on their analysis, we  obtain
\begin{equation}{R_{cmb}} \le h{R_C},\end{equation}
where $R_{cmb}$ denotes the centralized regret with mini-batch and $h$ is the size of  $H_t$. 

\begin{figure}[tbp]
 \subfigure [Synthetic data with nodes=64]{   
 \label{fig:subfig:a}
\includegraphics[width=1.7in]{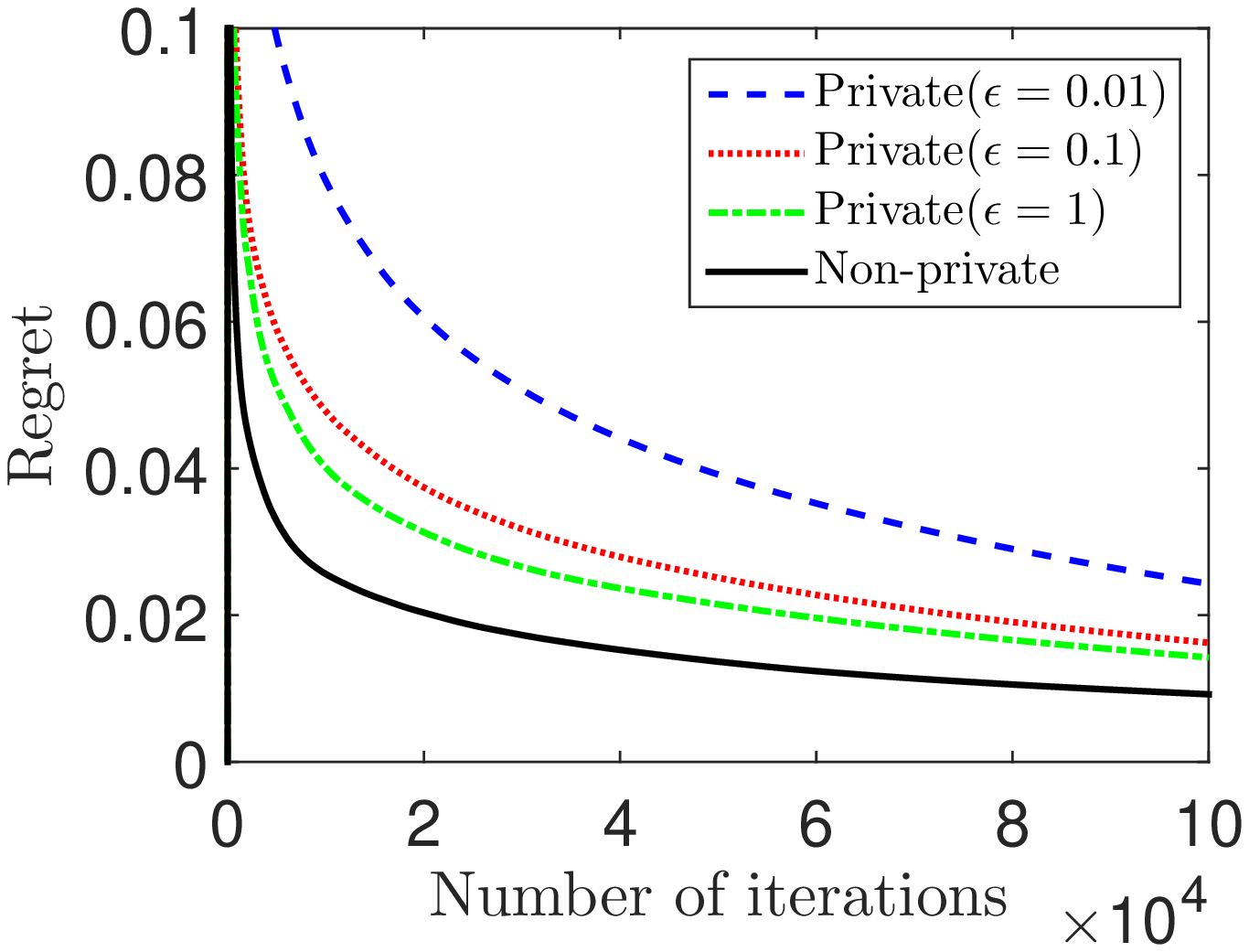}} 
\hspace{0.001in}
 \subfigure[RCV1 data with nodes=64]{   
  \label{fig:subfig:b} 
  \includegraphics[width=1.7in]{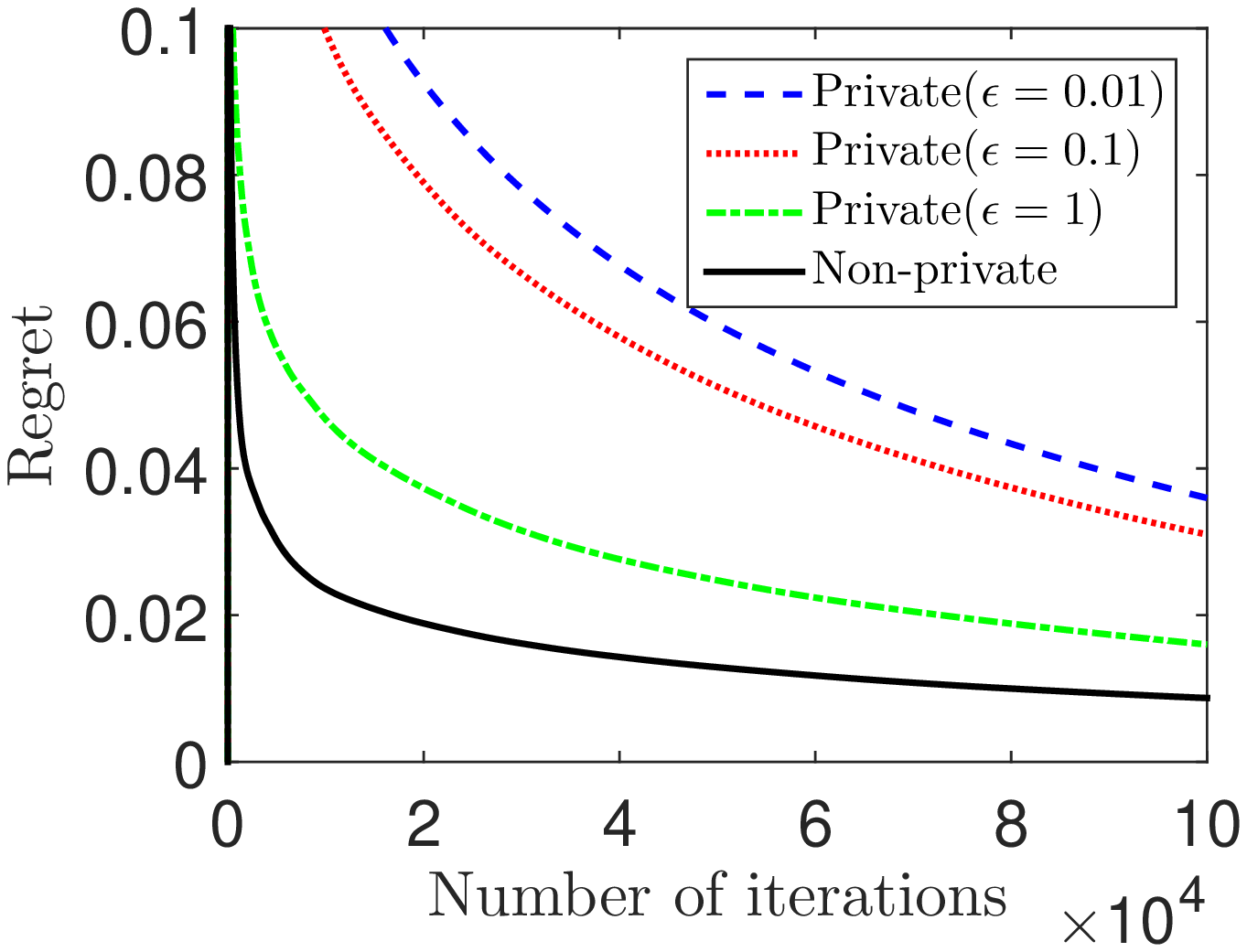}} 
\subfigure[Synthetic data with $\epsilon=0.1$]{   
  \label{fig:subfig:c} 
  \includegraphics[width=1.7in]{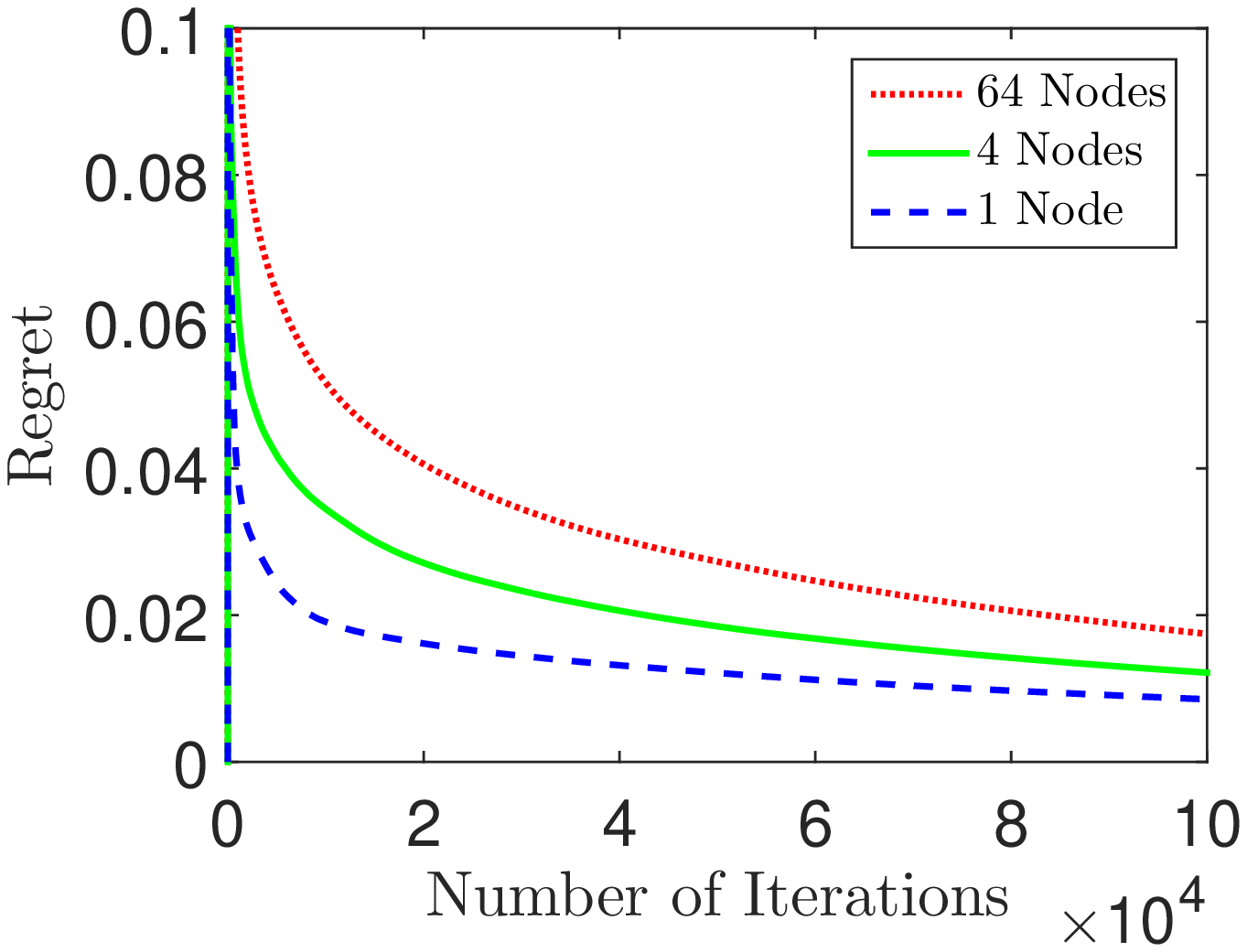}} 
\hspace{0.001in}
\subfigure[RCV1 data with $\epsilon=0.1$]{   
  \label{fig:subfig:d} 
  \includegraphics[width=1.7in]{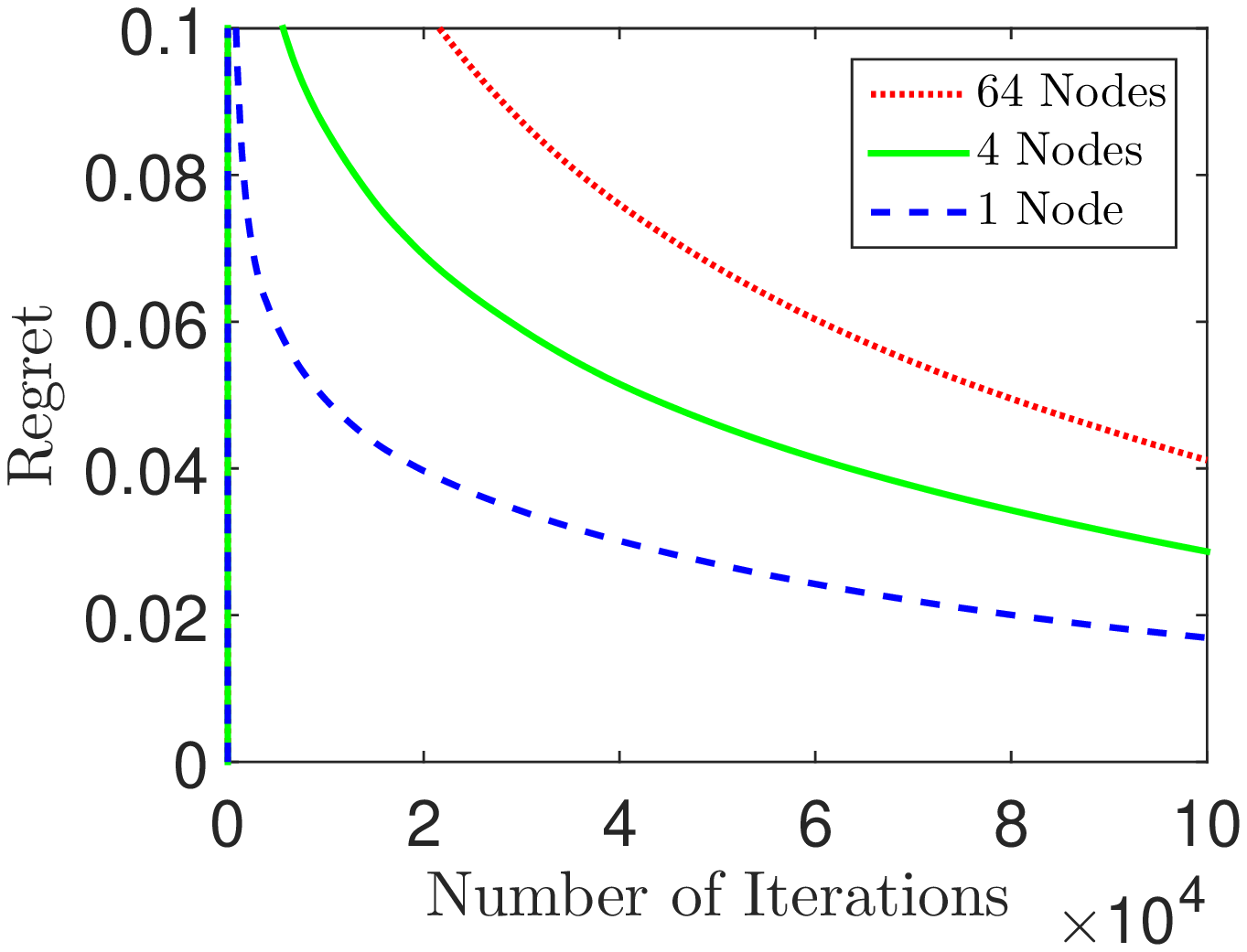}} 
 \caption{ (a) and (b): Regret vs Privacy on synthetic and RCV1 datasets.  (c) and (d): Regret vs Nodes on synthetic and  RCV1  datasets. Note that the y-axis denotes the  average regret (normalized by the number of iterations).
} 
 \label{fig:subfig} 
\end{figure}

\textbf{ Lemma 10.} Under Assumption 1, for the centralized offline learning update with mini-batch, if we update the iterate as (55), then with probability  $1 -4 \gamma \ln T$, we have
\begin{align}
&\notag{F_{mb}}\left( {\overline w } \right) - {F_{mb}}\left( {{w^ * }} \right)\\
&\notag \le \frac{{{h^2}{R_C}}}{T} + 4\sqrt {\frac{{{L^2}\ln \left( {{T \mathord{\left/
 {\vphantom {T h}} \right.
 \kern-\nulldelimiterspace} h}} \right)}}{\lambda }} \frac{{h\sqrt {h{R_C}} }}{T}\\
&\quad\, + \max \left\{ {\frac{{16{L^2}}}{\lambda },6} \right\}\frac{{h\ln \left( {{1 \mathord{\left/
 {\vphantom {1 \gamma }} \right.
 \kern-\nulldelimiterspace} \gamma }} \right)}}{T}
\end{align}

\begin{proof}
Substituting $T/h$ (see step 2 in Algorithm 2) for $T$ in (58) and using ${R_{cmb}} \le h{R_C}$, we  obtain (60).
\end{proof}

Lemma 10 is the  utility analysis for the centralized model, while Algorithm 2 is a distributed offline learning algorithm using mini-batch. Next, we analyze the utility of the distributed model on the basis of  Lemma 10.  Similarly, we shall use the regret of  Algorithm 1 to achieve the fast convergence rate for Algorithm 2. 

\textbf{Theorem 4 (Utility of Algorithm 2).} Under Assumption 1, the regret $R_D$  of Algorithm 1 can be used to achieve the convergence rate for Algorithm 2. Then, with  probability  $1 -4 \gamma \ln T$, we have
\begin{align}
&\notag{F_{dmb}}\left( {{{\overline w }^i}} \right) - {F_{dmb}}\left( {{w^i}^*} \right)\\
&\notag \le \frac{{{h^2}{R_D}}}{{mT}} + 4\sqrt {\frac{{{L^2}\ln \left( {{T \mathord{\left/
 {\vphantom {T h}} \right.
 \kern-\nulldelimiterspace} h}} \right)}}{\lambda }} \frac{{h\sqrt {{{h{R_D}} \mathord{\left/
 {\vphantom {{h{R_D}} m}} \right.
 \kern-\nulldelimiterspace} m}} }}{T}\\
& \quad\,+ \max \left\{ {\frac{{16{L^2}}}{\lambda },6} \right\}\frac{{h\ln \left( {{1 \mathord{\left/
 {\vphantom {1 \gamma }} \right.
 \kern-\nulldelimiterspace} \gamma }} \right)}}{T},
\end{align}
where $F_{dmb}\left( w \right) = E_{dmb}\left[ {f\left( {w^i,x,y} \right)} \right]$, ${\overline w ^i} = \frac{1}{{{T \mathord{\left/
 {\vphantom {T h}} \right. \kern-\nulldelimiterspace} h}}}\sum\limits_{t = 1}^{{T \mathord{\left/ {\vphantom {T h}} \right.
 \kern-\nulldelimiterspace} h}} {w_t^i}$.

\begin{proof}
We  estimate the convergence rate  with respect to an  arbitrary learner $i$. So we use the regret of a single learner, $R_D/m$. Based on  (60), we substitute  $R_D/m$  for $R_C$, then obtain (61).
\end{proof}

Based on \cite{jain2011differentially} and \cite{kakade2009generalization}, we study the application of regret bounds to offline convergence rates in distributed  scenarios. Our work also have the same three significant advantages in \cite{jain2011differentially}. Except for these existing advantages, we find  new advantages in distributed scenarios: 1) the corresponding algorithms converge faster; 2) guaranteeing the same level of privacy needs less noise; 3) the noise of same magnitude  has less influence on the utility of algorithms.

\begin{figure}[tbp]
 \subfigure[Synthetic data with size=1]{   
 \label{fig:subfig:s1} 
\includegraphics[width=1.7in]{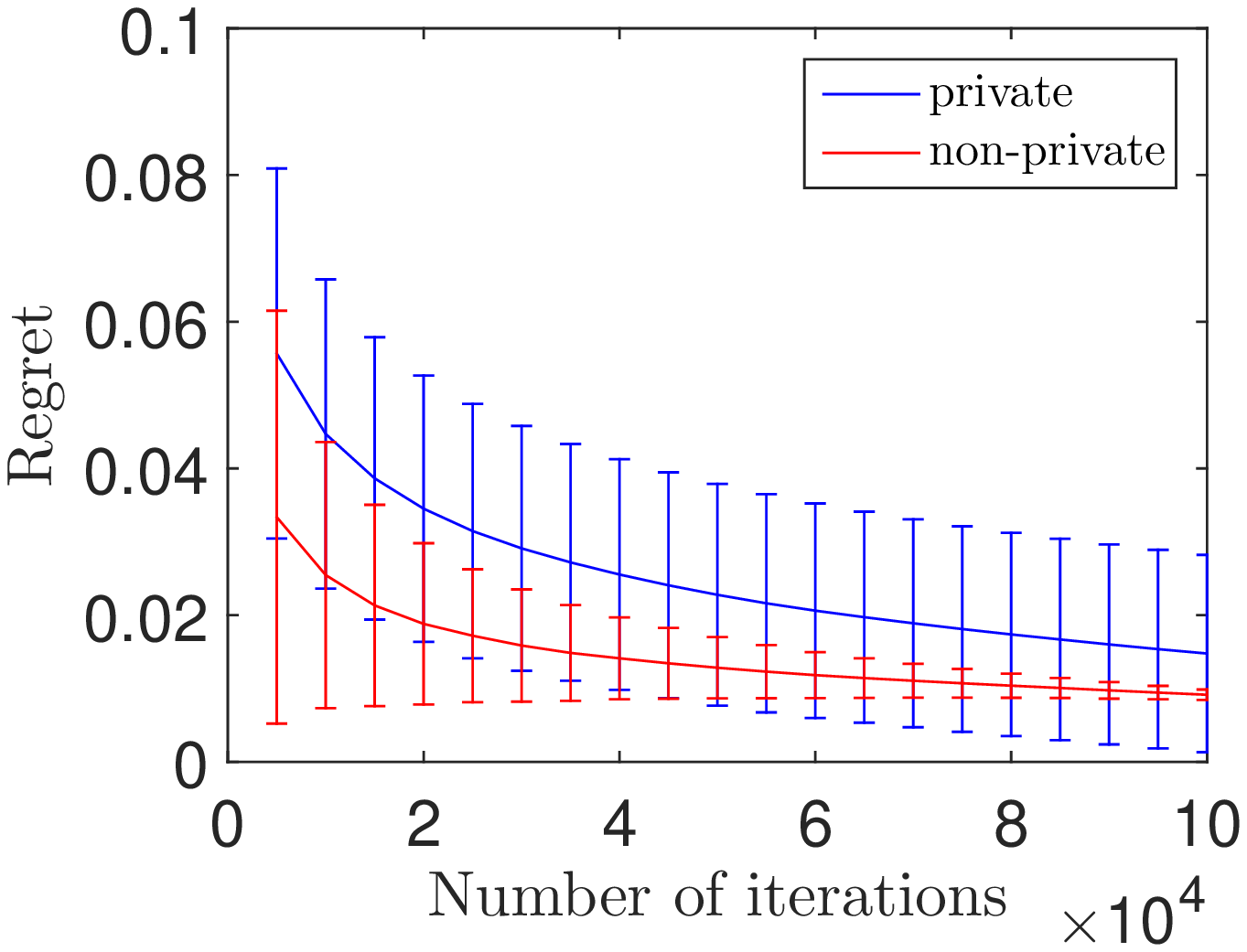}} 
\hspace{0.001in}
 \subfigure[Synthetic data with size=5]{   
  \label{fig:subfig:s5} 
  \includegraphics[width=1.7in]{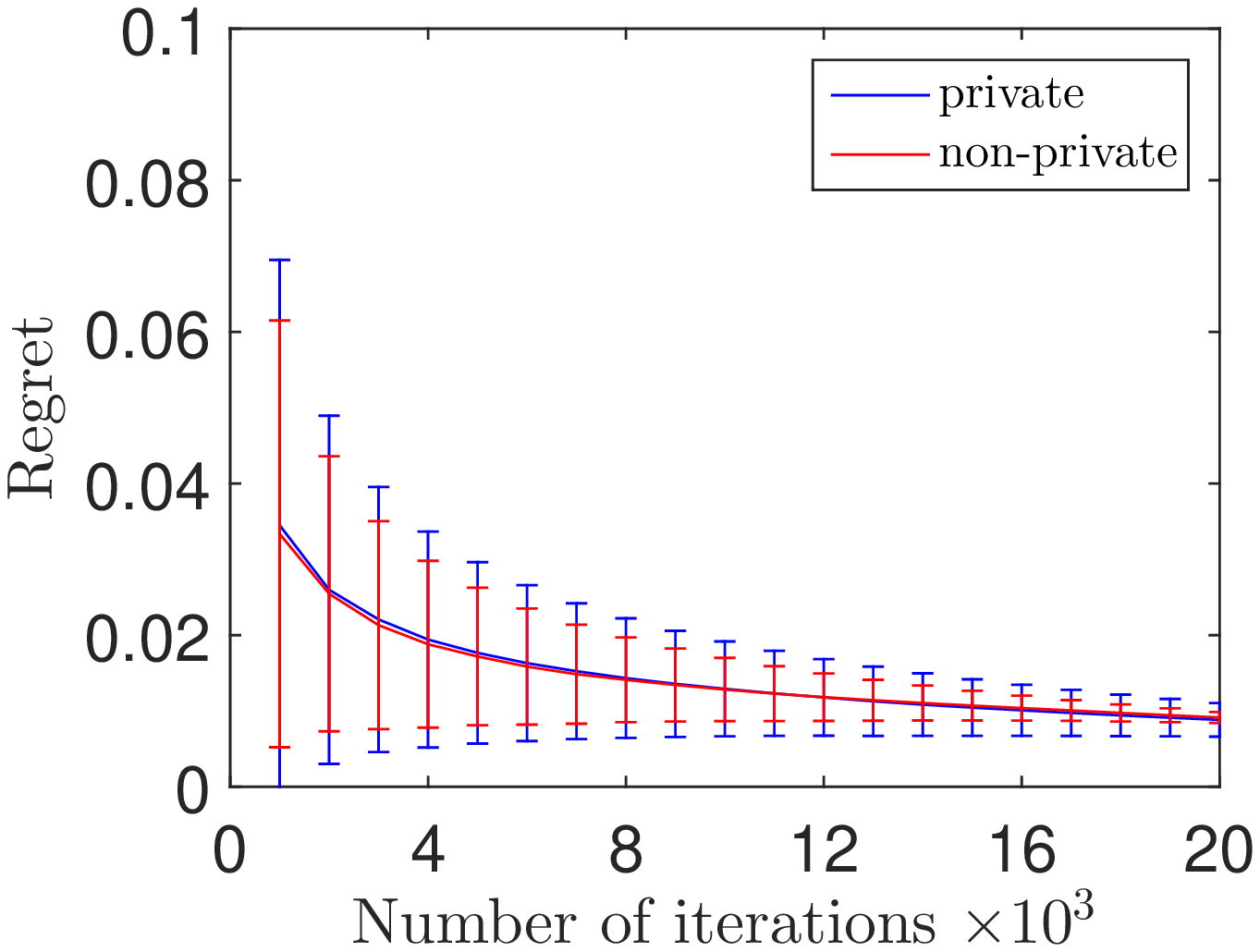}} 
\subfigure[RCV1 data with size=1]{   
  \label{fig:subfig:u1} 
  \includegraphics[width=1.7in]{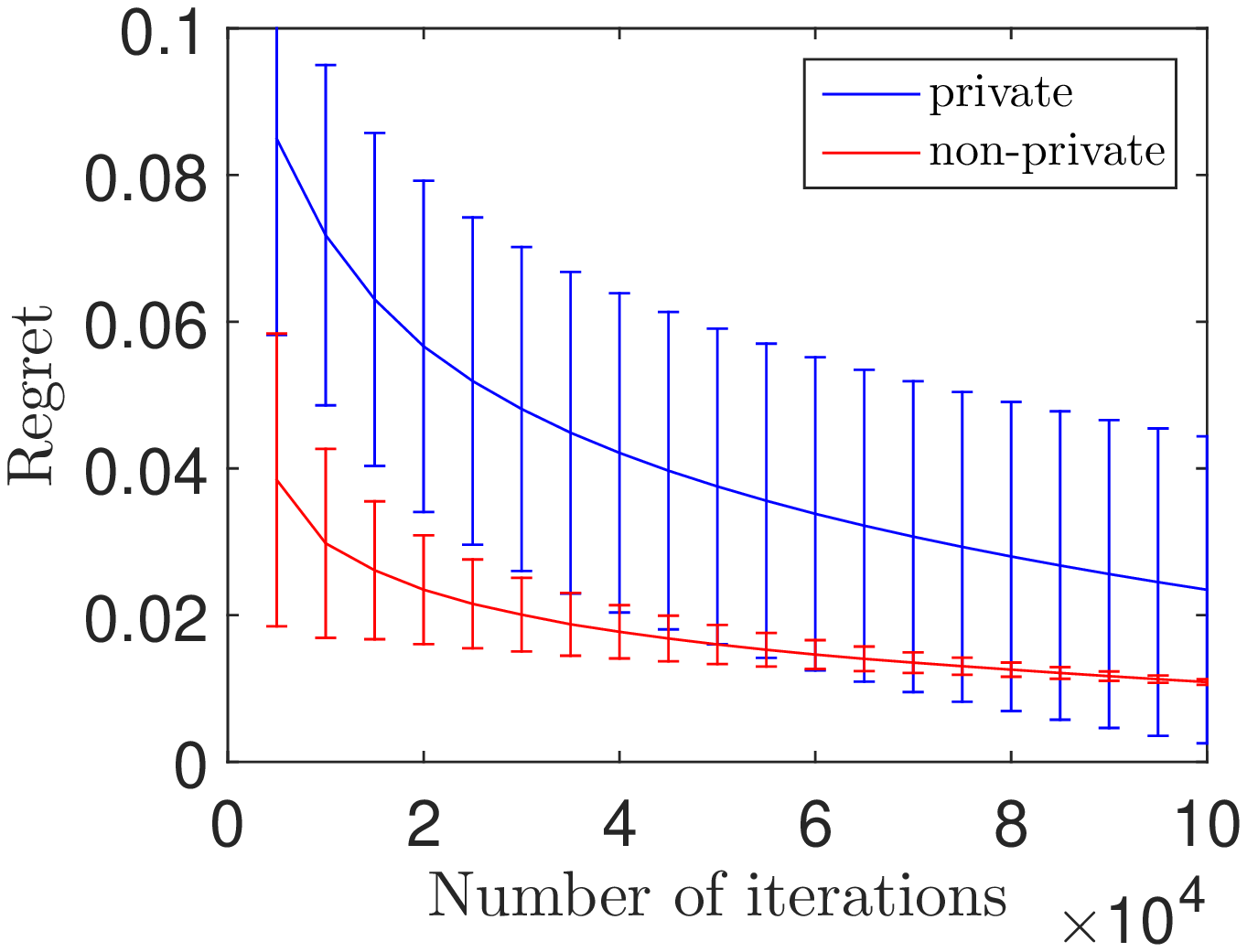}} 
\hspace{0.001in}
\subfigure[RCV1 data with size=5]{   
  \label{fig:subfig:u5} 
  \includegraphics[width=1.7in]{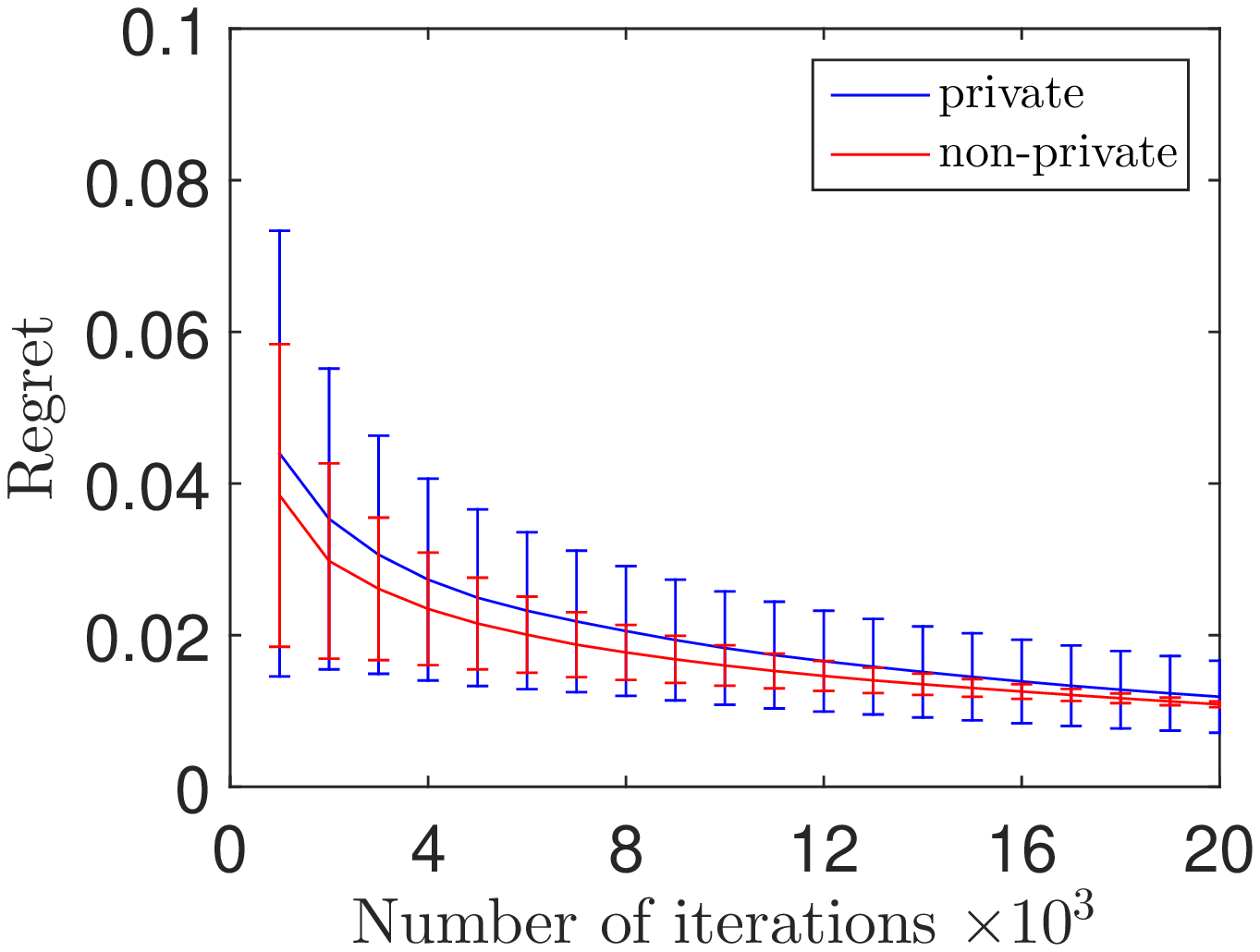}} 
 \caption{ (a) and (b): Regret vs Batch size on synthetic dataset. (c) and (d): Regret vs Batch size on RCV1  dataset. Note that this figure shows the variance and mean of  the  average regret (normalized by the number of iterations).
} 
 \label{fig:subfig} 
\end{figure}

\section{simulations}
In this section, we conduct two sets  of simulations. One is to study the privacy and regret trade-offs for our DOLA. The other is to illustrate how well the mini-batch  performs to reduce high variance of differential privacy in the offline learning algorithm. For our implementations, we have  the hinge loss function $f_t^i\left( w \right) = \max \left( {1 - y_t^i\left\langle {w,x_t^i} \right\rangle } \right)$, where $\left\{ {\left( {x_t^i,y_t^i} \right) \in {\mathbb{R}^n} \times \left\{ { \pm 1} \right\}} \right\}$ are the data available only to the $i$-th learner. For fast convergence rates, we set the learning rate ${\alpha _t} = \frac{1}{{\lambda t}}$. Furthermore, we  do experiments  on both  synthetic and real datasets. The synthetic data are generated from a unit ball of dimensionality $d=10$. We generate a total of 100,000 labeled examples. The real data used in our simulation  is a subset of the RCV1 dataset. For a  sharp contrast,  this subset has  the same number of examples  with the synthetic data. As shown in Algorithm 1 and 2, the dataset is divided into $m$ subsets. Each node updates the parameter based on its own subset and timely exchanges the updates its parameter to neighbors. Note that at round $t$, the $i$-th learner  must exchange the parameter $w_t^i$ in strict accordance with Assumption 2. For a good observation, we sum the normalized error bounds (i.e., the ``Regret'' on y-axis) for both Figure 1 and 2.

Figure 1 (a) and (b) show the average regret (normalized by the number of iterations) incurred by our DOLA for different level of privacy $\epsilon$ on synthetic and RCV1 datasets. Our differentially private DOLA  has low-regret even for a little high level of privacy (e.g., $\epsilon=0.01$). The regret obtained by the non-private algorithm has the lowest regret as expected. More significantly,  the  regret gets closer to the non-private regret as its privacy preservation is weaker. Figure 1 (c) and (d) show the  average regret for different nodes of  the online system on the same level of privacy. Clearly,  the centralized online learning algorithm ($node=1$) has the lowest regret  on the level of privacy $\epsilon=0.1$ and the regret gets lower as its number of nodes is  smaller.  Furthermore, the regret on synthetic data performs better than that on real data under the same conditions.

Figure 2 (a) and (b) show the average regret for different batch size on synthetic data. When batch size is one (see Figure 2 (a)), the differentially private regret has higher variance than the non-private regret. However, a modest batch size $h=5$, as shown in  Figure 2 (b),  reduces  the variance of our differentially private distributed offline learning algorithm. The mini-batch technique makes the variance of differentially private distributed offline learning algorithm nearly identical to that of  the non-private offline algorithm. Figure 2 (c) and (d) show the same simulation on RCV1 data  and obtain the same conclusion with Figure 2 (a) and (b).
\begin{table}[htbp]
\centering \caption{\label{tab:test}}
\begin{tabular}{|*{4}{c|}}
\hline
\multicolumn{2}{|c|}{Method} & Nodes & Accuracy\\
\hline
\multicolumn{2}{|c|}{\multirow{3}{*}{Non-private}}&$1$& $82.51\%$\\
\cline{3-4} 
\multicolumn{2}{|c|}{}&$4$ &$74.64\%$ \\
\cline{3-4} 
\multicolumn{2}{|c|}{}&$64$&$65.72\%$\\
\hline
\multirow{9}{*}{Private}
& $\epsilon=1$ & $1$ &  $82.51\%$\\\cline{2-4} 
&$\epsilon=1$ & $4$& $74.64\%$\\\cline{2-4}  
&$\epsilon=1$ & $64$ &$65.72\%$\\\cline{2-4}  
&$\epsilon=0.1$ &  $1$& $80.17\%$\\\cline{2-4} 
&$\epsilon=0.1$ & $4$& $70.86\%$\\\cline{2-4} 
&$\epsilon=0.1$ & $64$& $62.34\%$\\\cline{2-4} 
&$\epsilon=0.01$ &  $1$& $75.69\%$\\\cline{2-4} 
&$\epsilon=0.01$ &  $4$& $64.81\%$\\\cline{2-4}  
&$\epsilon=0.01$ & $64$& $50.36\%$\\
\hline
\end{tabular}
\end{table}

As we know, the hinge loss $\ell \left( {w} \right) = \max \left( {1 - y{w^{\rm T}}x,0} \right)$  leads to the  data mining algorithm,  SVM. To be more persuasive, we conduct a differentially private distributed SVM and test this algorithm on RCV1 data. Table 1 shows the accuracy for different level of privacy and different number of nodes  of algorithm. Intuitively, the centralized non-private model has the highest accuracy $88.74\%$ while the model of $64$ nodes at a high level  $\epsilon=0.01$ of privacy has the lowest accuracy $50.36\%$.
 Further, we conclude that the accuracy gets higher as the level of privacy is lower or the number of nodes is smaller. This conclusion goes  along with Figure 1 and 2.

\section{conclusion and discussion}
 We have proposed a differentially private  distributed online learning algorithm. We used subgradient to update the learning parameter and used random doubly stochastic matrix to guide the learners to communicate with others. More importantly, our network topology is time-variant. As expected, we obtained the regret bounds in the order of $O(\sqrt T )$ and $O(\log T)$. Interestingly, the magnitude of the  total noise added to guarantee  $\epsilon$-differential privacy also has the order of  $O(\sqrt T )$ and $O(\log T)$ along with the non-private regret.

Furthermore, we used our private distributed online learning algorithm with good regret bounds to solve the private distributed offline learning problems. In order to reduce high variance of our differentially private algorithm, we use the mini-batch technique to weaken the influence of added noise. This method  makes the algorithm guarantee the same level of privacy using less random noise.

In this paper, we did not take the delay into  consideration. In distributed online learning scenarios, there must exist delays among the nodes when they communicate with others, which  is hard to analyze.  Because each node has different delay according to its communication graph and the  graph is even time-variant.  Then, in future work, we hope that  distributed online learning with delay can be presented.

\ifCLASSOPTIONcompsoc
  \section*{Acknowledgments}
\else
  \section*{Acknowledgment}
\fi
This research is supported by National Science Foundation of China with Grant 61401169.

\bibliographystyle{IEEEtran}
\bibliography{test.bib}

\begin{thebibliography}{10}
\providecommand{\url}[1]{#1}
\csname url@samestyle\endcsname
\providecommand{\newblock}{\relax}
\providecommand{\bibinfo}[2]{#2}
\providecommand{\BIBentrySTDinterwordspacing}{\spaceskip=0pt\relax}
\providecommand{\BIBentryALTinterwordstretchfactor}{4}
\providecommand{\BIBentryALTinterwordspacing}{\spaceskip=\fontdimen2\font plus
\BIBentryALTinterwordstretchfactor\fontdimen3\font minus
  \fontdimen4\font\relax}
\providecommand{\BIBforeignlanguage}[2]{{%
\expandafter\ifx\csname l@#1\endcsname\relax
\typeout{** WARNING: IEEEtran.bst: No hyphenation pattern has been}%
\typeout{** loaded for the language `#1'. Using the pattern for}%
\typeout{** the default language instead.}%
\else
\language=\csname l@#1\endcsname
\fi
#2}}
\providecommand{\BIBdecl}{\relax}
\BIBdecl

\bibitem{dwork2006differential}
C.~Dwork, ``Differential privacy,'' in \emph{Proceedings of the 33rd
  international conference on Automata, Languages and Programming-Volume Part
  II}.\hskip 1em plus 0.5em minus 0.4em\relax Springer-Verlag, 2006, pp. 1--12.

\bibitem{olfati2007}
R.~Olfati-Saber, J.~Fax, and R.~Murray, ``Consensus and cooperation in
  networked multi-agent systems,'' \emph{Proceedings of the IEEE}, vol.~95,
  no.~1, pp. 215--233, Jan 2007.

\bibitem{ram2010distributed}
S.~Ram, A.~Nedi{\'c}, and V.~Veeravalli, ``Distributed stochastic subgradient
  projection algorithms for convex optimization,'' \emph{Journal of
  optimization theory and applications}, vol. 147, no.~3, pp. 516--545, 2010.

\bibitem{nedic2009distributed}
A.~Nedic and A.~Ozdaglar, ``Distributed subgradient methods for multi-agent
  optimization,'' \emph{IEEE Transactions on Automatic Control}, vol.~54,
  no.~1, pp. 48--61, 2009.

\bibitem{yuan2013convergence}
K.~Yuan, Q.~Ling, and W.~Yin, ``On the convergence of decentralized gradient
  descent,'' \emph{arXiv preprint arXiv:1310.7063}, 2013.

\bibitem{Yan2013}
F.~Yan, S.~Sundaram, S.~Vishwanathan, and Y.~Qi, ``Distributed autonomous
  online learning: Regrets and intrinsic privacy-preserving properties,''
  \emph{IEEE Transactions on Knowledge and Data Engineering}, vol.~25, no.~11,
  pp. 2483--2493, 2013.

\bibitem{jain2011differentially}
P.~Jain, P.~Kothari, and A.~Thakurta, ``Differentially private online
  learning,'' \emph{arXiv preprint arXiv:1109.0105}, 2011.

\bibitem{chaudhuri2011}
K.~Chaudhuri, C.~Monteleoni, and A.~D. Sarwate, ``Differentially private
  empirical risk minimization,'' \emph{The Journal of Machine Learning
  Research}, vol.~12, pp. 1069--1109, 2011.

\bibitem{williams2010}
O.~Williams and F.~McSherry, ``Probabilistic inference and differential
  privacy,'' in \emph{Advances in Neural Information Processing Systems}, 2010,
  pp. 2451--2459.

\bibitem{kakade2009generalization}
S.~M. Kakade and A.~Tewari, ``On the generalization ability of online strongly
  convex programming algorithms,'' in \emph{Advances in Neural Information
  Processing Systems}, 2009, pp. 801--808.

\bibitem{song2013stochastic}
S.~Song, K.~Chaudhuri, and A.~D. Sarwate, ``Stochastic gradient descent with
  differentially private updates,'' in \emph{IEEE Global Conference on Signal
  and Information Processing}, 2013.

\bibitem{zinkevich2003}
M.~Zinkevich, ``Online convex programming and generalized infinitesimal
  gradient ascent,'' \emph{In ICML}, pp. 928--936, 2003.

\bibitem{hazan2007}
E.~Hazan, A.~Agarwal, and S.~Kale, ``Logarithmic regret algorithms for online
  convex optimization,'' \emph{Machine Learning}, vol.~69, no. 2-3, pp.
  169--192, 2007.

\bibitem{huang2015}
Z.~Huang, S.~Mitra, and N.~Vaidya, ``Differentially private distributed
  optimization,'' in \emph{Proceedings of the 2015 International Conference on
  Distributed Computing and Networking}.\hskip 1em plus 0.5em minus 0.4em\relax
  ACM, 2015, p.~4.

\bibitem{hazan2015online}
E.~Hazan, ``\href{http://ocobook.cs.princeton.edu/OCObook.pdf}{Online Convex
  Optimization},'' 2015.

\bibitem{duchi2012dual}
J.~C. Duchi, A.~Agarwal, and M.~J. Wainwright, ``Dual averaging for distributed
  optimization,'' in \emph{Communication, Control, and Computing (Allerton),
  2012 50th Annual Allerton Conference on}.\hskip 1em plus 0.5em minus
  0.4em\relax IEEE, 2012, pp. 1564--1565.

\bibitem{rajkumar2012}
A.~Rajkumar and S.~Agarwal, ``A differentially private stochastic gradient
  descent algorithm for multiparty classification,'' in \emph{International
  Conference on Artificial Intelligence and Statistics}, 2012, pp. 933--941.

\bibitem{Bassily2014}
R.~Bassily, A.~Smith, and A.~Thakurta, ``Differentially private empirical risk
  minimization: Efficient algorithms and tight error bounds,'' \emph{arXiv
  preprint arXiv:1405.7085}, 2015.

\bibitem{mcsherry2009}
F.~D. McSherry, ``Privacy integrated queries: an extensible platform for
  privacy-preserving data analysis,'' in \emph{Proceedings of the 2009 ACM
  SIGMOD International Conference on Management of data}.\hskip 1em plus 0.5em
  minus 0.4em\relax ACM, 2009, pp. 19--30.

\bibitem{dekel2012optimal}
O.~Dekel, R.~Gilad-Bachrach, O.~Shamir, and L.~Xiao, ``Optimal distributed
  online prediction using mini-batches,'' \emph{The Journal of Machine Learning
  Research}, vol.~13, no.~1, pp. 165--202, 2012.

\end{thebibliography}

\begin{IEEEbiography}[{\includegraphics[width=1in,height=1.25in,clip,keepaspectratio]{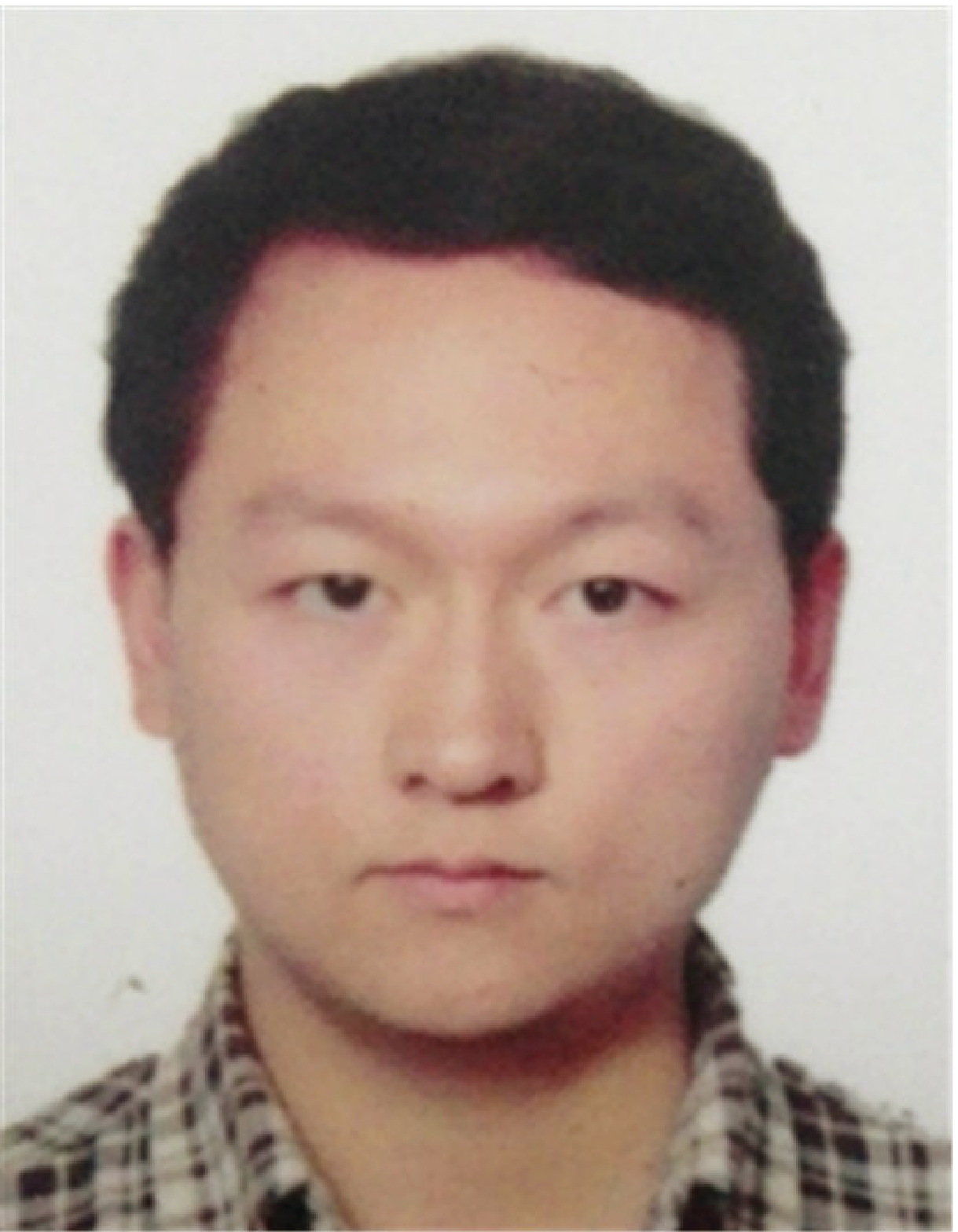}}]
{Chencheng Li (S'15)} received the B.S. degree from
Huazhong University of Science and Technology,
Wuhan, P.R. China, in 2014 and is currently
working toward the M.S. degree at the School
of Electronic Information and Communications,
Huazhong University of Science and
Technology, Wuhan, P.R. China. His current
research interest includes:  online learning in Big Data and differential privacy. He is a student member
of the IEEE.
\end{IEEEbiography}

\begin{IEEEbiography}[{\includegraphics[width=1in,height=1.25in,clip,keepaspectratio]{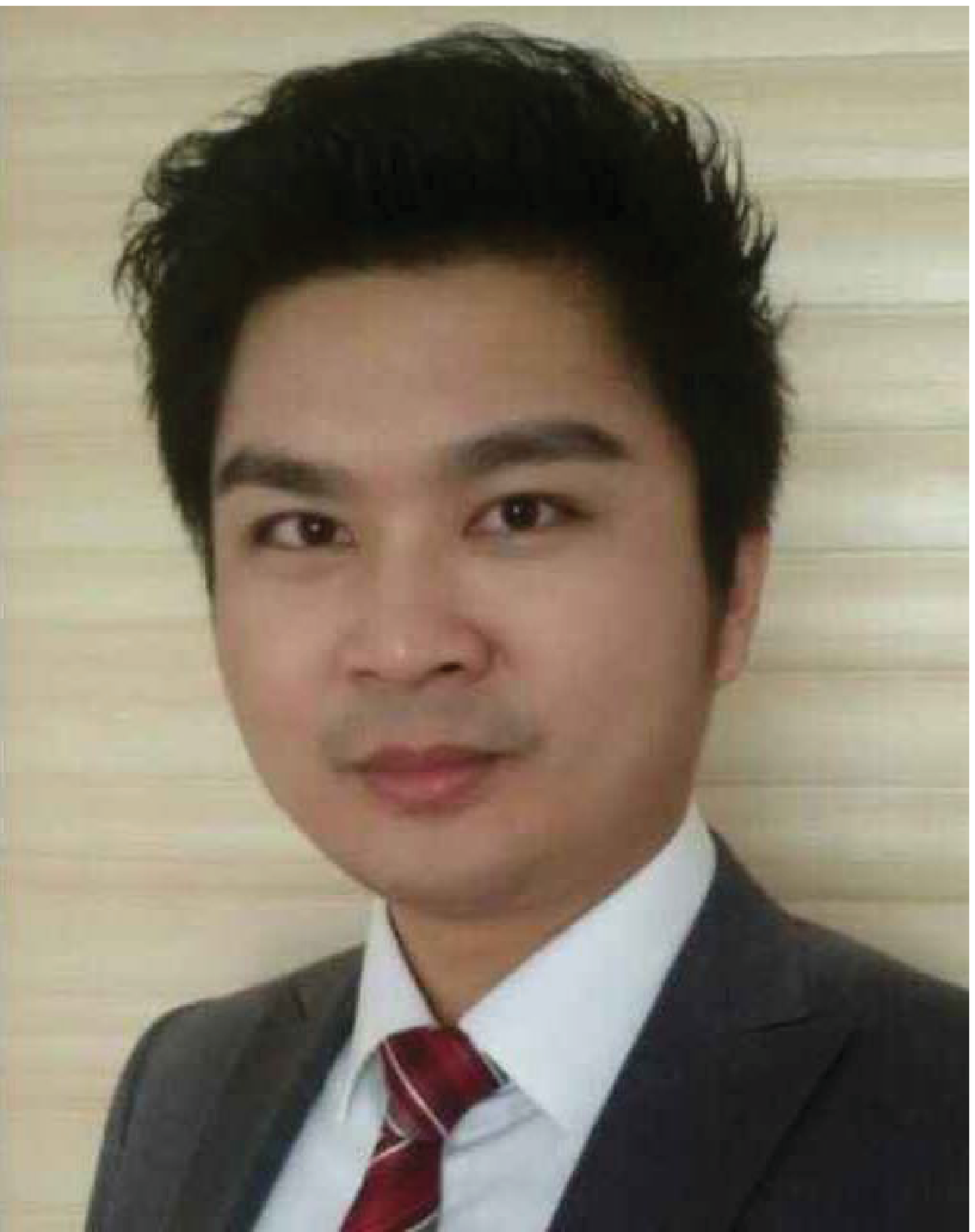}}]{Pan
Zhou (S'07--M'14)} is currently an associate professor
 with School of Electronic Information and Communications, Huazhong
University of Science and Technology, Wuhan, P.R. China. He received
his Ph.D. in the School of Electrical and Computer Engineering at the
Georgia
Institute of Technology (Georgia Tech) in 2011, Atlanta, USA. He
received his B.S. degree in the \emph{Advanced Class} of
HUST, and a M.S. degree in the Department of Electronics and
Information Engineering
from HUST, Wuhan, China, in 2006 and 2008, respectively.
He held honorary degree in his bachelor and merit research award
of HUST in his master study. He was a
senior technical memeber at Oracle Inc, America during 2011 to 2013,
Boston, MA, USA,  and worked on hadoop and distributed storage system
for big data
analytics at Oralce cloud Platform.  His current research interest
includes:  communication and information networks, security and
privacy,  machine learning and big data.
\end{IEEEbiography}

\begin{IEEEbiography}[{\includegraphics[width=1in,height=1.25in,clip,keepaspectratio]{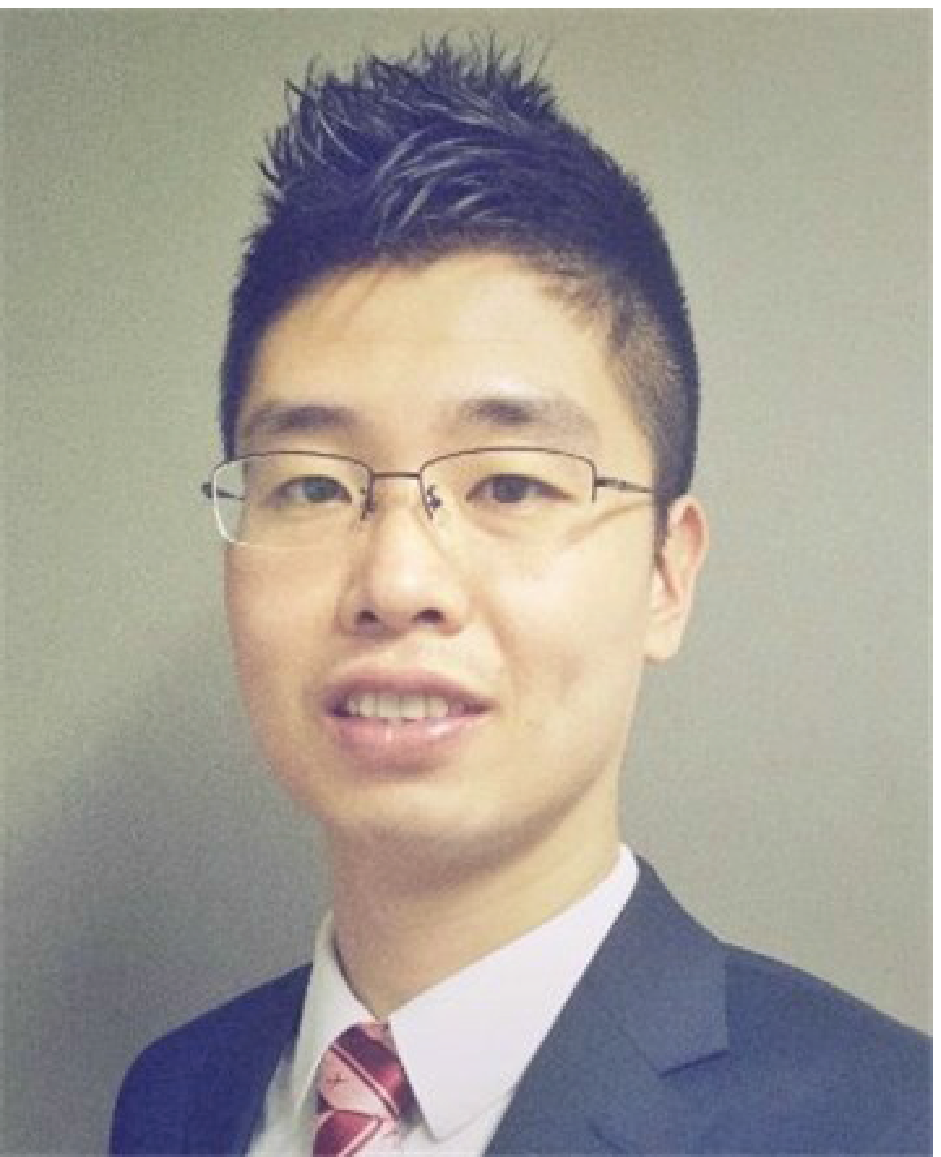}}]{
Gong Chen (M'12)} received his B.S. degree in Electronic Science and Technology
from Beijing University of Posts and Telecommunications, China in 2007
and his Dipl\^{o}me d'Ing\'{e}nieur in Electronics and Signal
Processing from INP-ENSEEIHT, France in 2010 and M.S. in Electrical
and Computer Engineering from Georgia Tech, GA in 2011 and  is currently
working toward the Ph.D. degree  in ECE at Georgia Tech.
 He works with Dr. John Copeland and his  current
research interest is improving security for digital advertising
ecosystems.
\end{IEEEbiography}

\begin{IEEEbiographynophoto}{Tao Jiang (M'06--SM'10)}
 is currently a full Professor in the School of Electronic Information
and Communications, Huazhong University of Science and Technology,
Wuhan, P. R. China. He received the B.S. and M.S. degrees in applied
geophysics from China University of Geosciences, Wuhan, P. R. China,
in 1997 and 2000, respectively, and the Ph.D. degree in information
and communication engineering from Huazhong University of Science and
Technology, Wuhan, P. R. China, in April 2004. From Aug. 2004 to Dec.
2007, he worked in some universities, such as Brunel University and
University of Michigan-Dearborn, respectively. He has authored or
co-authored over 160 technical papers in major journals and
conferences and six books/chapters in the areas of communications and
committee membership of some major IEEE conferences, including
networks. He served or is serving as symposium technical program
INFOCOM, GLOBECOM, and ICC, etc.. He is invited to serve as TPC
Symposium Chair for the IEEE GLOBECOM 2013 and IEEEE WCNC 2013. He is
served or serving as associate editor of some technical journals in
communications, including in IEEE Communications Surveys and
Tutorials, IEEE Transactions on Vehicular Technology, and IEEE
Internet of Things Journal, etc.. He is a recipient of the NSFC for
Distinguished Young Scholars Award in P. R. China. He is a senior
member of IEEE.
\end{IEEEbiographynophoto}

\end{document}